\documentclass{article}

\PassOptionsToPackage{numbers, compress}{natbib}

\usepackage[preprint]{neurips_2026}

\usepackage[utf8]{inputenc} % allow utf-8 input
\usepackage[T1]{fontenc}% use 8-bit T1 fonts
\usepackage{hyperref}% hyperlinks
\usepackage{url} % simple URL typesetting
\usepackage{booktabs}% professional-quality tables
\usepackage{amsfonts}% blackboard math symbols
\usepackage{nicefrac}% compact symbols for 1/2, etc.
\usepackage{microtype}  % microtypography
\usepackage{xcolor}  % colors

\usepackage{multicol}
\usepackage{multirow}
\usepackage{graphicx}
\usepackage{subcaption}
\usepackage{adjustbox} 
\usepackage{wrapfig}
\usepackage{amsmath}
\usepackage{algorithm}
\usepackage{algorithmic}

\hypersetup{hidelinks}
\newcommand{\bc}[1]{#1}
\title{Keep Rehearsing and Refining: Lifelong Learning Vehicle Routing under Continually Drifting Tasks}

\author{%
  Jiyuan Pei \\
  Center of Data Science and \\ Artificial Intelligence  \& School of \\ Engineering and Computer Science\\
  Victoria University of Wellington\\
  \texttt{jiyuan.pei@vuw.ac.nz} \\
  \And
  Yi Mei\\
  Center of Data Science and \\ Artificial Intelligence  \& School of \\ Engineering and Computer Science\\
  Victoria University of Wellington\\
  \texttt{yi.mei@ecs.vuw.ac.nz} \\
  \And
  Jialin Liu\\
  School of Data Science \\ 
  Lingnan University\\
  \texttt{jialin.liu@ln.edu.hk} \\
  \And
  Mengjie Zhang\\
  Center of Data Science and \\ Artificial Intelligence  \& School of \\ Engineering and Computer Science\\
  Victoria University of Wellington\\
  \texttt{mengjie.zhang@ecs.vuw.ac.nz} \\
  \And
  Xin Yao\\
  School of Data Science \\ 
  Lingnan University\\
  \texttt{xinyao@ln.edu.hk} \\
}

\begin{document}

\maketitle

\begin{abstract}
  Existing neural solvers for vehicle routing problems (VRPs) are typically trained either in a one-off manner on a fixed set of pre-defined tasks or in a lifelong manner with tasks arriving sequentially, assuming sufficient training on each task. Both settings overlook a common real-world property: problem patterns may drift continually over time, yielding massive tasks sequentially arising, each with only limited training resources.
  % In this paper, we propose a novel lifelong learning paradigm for neural VRP solvers under continually drifting tasks over learning time steps, where at any time the training resource for any given task is insufficient.
  In this paper, we propose a novel lifelong learning paradigm for neural VRP solvers under continual task drift over time, where each task is locally stationary at one learning time step but receives only insufficient training resources.
  \bc{We empirically demonstrate that such continual drift arises in practice using a real-world logistics dataset.}
  We then propose \underline{\textbf{D}}ual \underline{\textbf{R}}eplay with \underline{\textbf{E}}xperience \underline{\textbf{E}}nhancement (DREE), a general framework to improve learning efficiency and mitigate catastrophic forgetting under such drift.
  Extensive experiments \bc{based on both the real-world logistics dataset and commonly used synthetic dataset} show that, under such continual drift, DREE effectively learns new tasks, preserves prior knowledge, improves generalization to unseen tasks, and can be applied to various existing neural solvers.
\end{abstract}

% Existing neural solvers for vehicle routing problems (VRPs) are typically trained either in a one-off manner on a fixed set of pre-defined tasks or in a lifelong manner with tasks arriving sequentially, assuming sufficient training on each task. Both settings overlook a common real-world property: problem patterns may drift continually over time, yielding massive tasks sequentially arising, each with only limited training resources. In this paper, we propose a novel lifelong learning paradigm for neural VRP solvers under continual task drift over time, where each task is locally stationary at one learning time step but receives only insufficient training resources. We empirically demonstrate that such continual drift arises in practice using a real-world logistics dataset. We then propose Dual Replay with Experience Enhancement (DREE), a general framework to improve learning efficiency and mitigate catastrophic forgetting under such drift. Extensive experiments based on both the real-world logistics dataset and commonly used synthetic dataset show that, under such continual drift, DREE effectively learns new tasks, preserves prior knowledge, improves generalization to unseen tasks, and can be applied to various existing neural solvers.

\section{Introduction}

Although neural solvers for vehicle routing problems (VRP) have achieved impressive performance by leveraging the strong capability of deep learning~\cite{regret,rethinkingNMOCO,huang2025rethinking}, especially deep reinforcement learning (DRL), they are predominantly built upon a restrictive assumption: abundant training resources are available from each task (problem pattern) that is pre-defined and stationary.
Training instances during learning are always independent and identically distributed, assuming patterns/tasks remain unchanged.

\begin{figure}[h] 
\centering 
  \begin{subfigure}{0.5\linewidth}
\centering
\includegraphics[width=\linewidth]{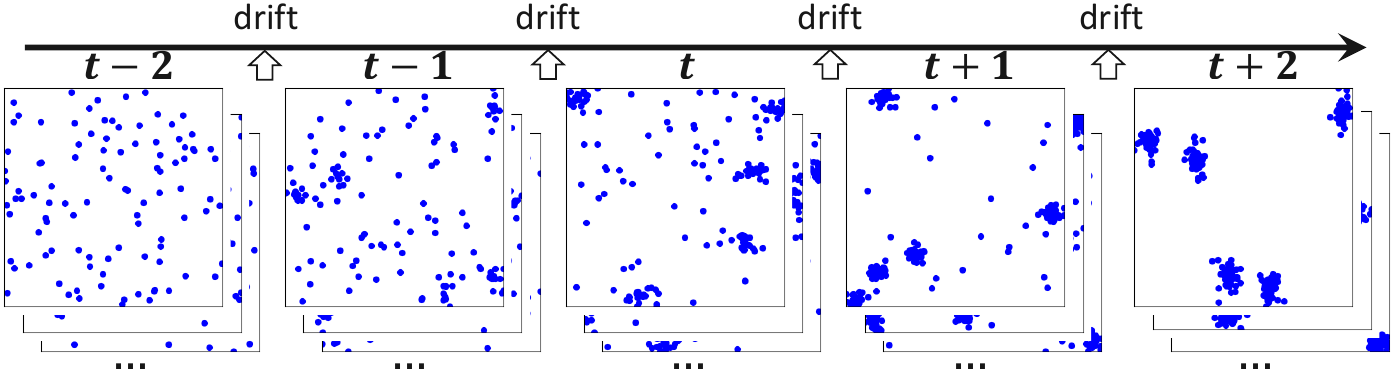}
\caption{Continually drifting tasks\label{fig:demo_ic}}
\label{fig:sub-b}
  \end{subfigure}

  \begin{subfigure}{0.5\linewidth}
\centering
\includegraphics[width=\linewidth]{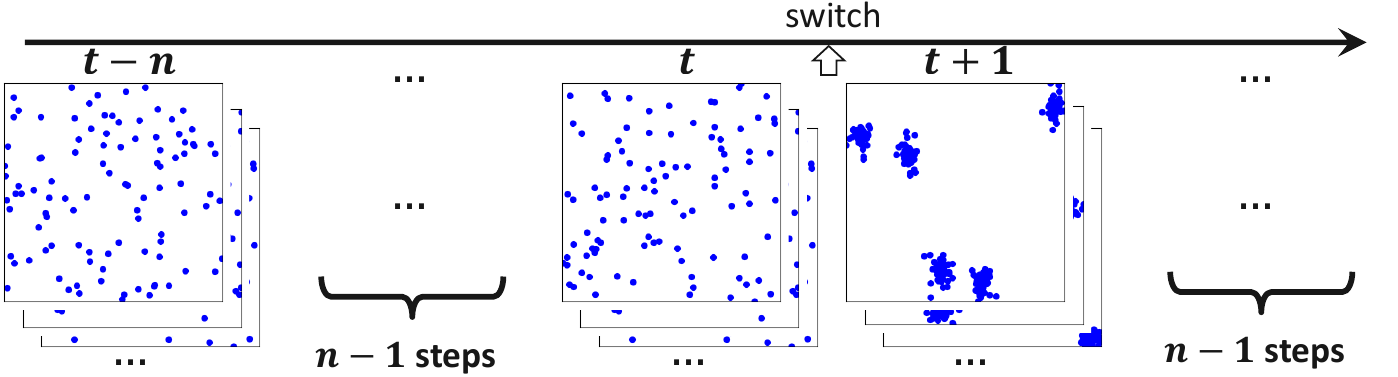}
% \vspace{-5pt} 
\caption{Periodically stationary tasks\label{fig:demo_sd}}
\label{fig:sub-a}
  \end{subfigure}

% \vspace{-4pt} 
  \caption{Lifelong learning of VRP, with node distribution changes from uniform to clustered. \label{fig:demo}}
\end{figure}

However, real-world scenarios often depart from this stationary regime~\cite{MARKOV2020104798,UCARPsurvey,EGPUCARP,math12010028,YTO}. New VRP instances arise sequentially over time, while their underlying patterns often drift continually (cf. Figure~\ref{fig:demo_ic}). 
\bc{An intuitive example is daily urban delivery, where a delivery service provider faces a continuous stream of new VRP instances generated by daily orders. The scale of orders and the underlying customer-location distributions may gradually change over time, as we observed (cf. Section \ref{sec:VPR_drift}) from the six months of delivery records in a comprehensive real-world logistics dataset~\cite{LaDe} covering five cities with populations ranging from millions to tens of millions. Such changes can be driven by factors such as business growth, shifting consumer trends, and the emergence or relocation of customer clusters.}
Under such drift, one-off training becomes insufficient, as solvers trained only on historical tasks tend to suffer performance degradation when facing newly emerging patterns~\cite{HistoricalDrift,HierarchicalDrfit}. \bc{However, simply fine-tuning the solver on each new pattern may lead to catastrophic forgetting of previously learned knowledge~\cite{surveyCL,surveyCRL}.}
This gap motivates the neural VRP solvers that can continue to learn from continually drifting tasks while maintaining competence on previously encountered tasks.

Some recent studies~\cite{li2024enhancing,feng2025lifelonglearner,LLR-BC} have begun exploring lifelong learning for neural VRP solvers, aiming to improve plasticity, i.e., the ability to learn new tasks, and stability, i.e., the ability to preserve knowledge of previously learned tasks. However, they mainly focus on the particular scenario considering periodically stationary tasks with sufficient per-task learning (denoted \textit{periodically stationary} scenario, cf. Figure~\ref{fig:demo_sd}). 
Learning of each task remains stationary for a long time interval (hundreds of training steps with millions of instances) before the next task.
Under this assumption, these methods typically rely on sufficient per-task training and on storing high-quality task-specific knowledge, such as model parameters~\cite{li2024enhancing,feng2025lifelonglearner} or behaviors~\cite{LLR-BC} to mitigate forgetting.
However, in real-world scenarios where the problem patterns drift continually, such assumptions no longer hold. Obtaining and storing high-quality per-task knowledge becomes impractical. As our experiments demonstrate (cf. Section~\ref{sec:performance}), these methods can suffer substantial performance degradation.

In this work, we propose a novel paradigm and provide a formal formulation of lifelong learning for neural VRP solvers, where the task changes continually during learning, with sufficient per-task training not available (referred to as \textit{continually drifting} scenario).
We propose \underline{\textbf{D}}ual \underline{\textbf{R}}eplay with \underline{\textbf{E}}xperience \underline{\textbf{E}}nhancement (DREE), a general and effective lifelong learning framework.
DREE buffers and replays encountered problem instances as well as the corresponding behaviors of the solver simultaneously (\textit{rehearsing}), so as to preserve rich knowledge under insufficient per-task training and mitigate forgetting caused by learning over a large variety of tasks. In addition, DREE proactively enhances the quality of buffered experiences through interactions between the two replay mechanisms (\textit{refining}), addressing the misleading issue from low-quality experiences induced by insufficient per-task training.
% \bc{Based on both widely used synthetic VRP tasks and real-world datasets,} comprehensive experimental studies on capacitated vehicle routing problems (CVRP) and traveling salesman problems (TSP) verify that DREE outperforms existing lifelong learning neural VRP methods in the continually drifting scenarios and has good applicability to different existing solvers.
\bc{We conduct extensive experiments on both synthetic continual-drift benchmarks, constructed from widely used instance-generation distributions, and a real-world logistics dataset.}
Experiment results verify that DREE outperforms all existing lifelong learning neural VRP methods and has good applicability to different existing solvers.

% Our main contributions are: (i) we introduce a novel paradigm and formal formulation of lifelong learning under continual task drift for neural VRP solvers, \bc{supported by analysis of such drift in a real-world logistics dataset}; (ii) we propose DREE, featuring novel dual replay and experience enhancement tailored for this paradigm; and (iii) we verify the effectiveness of DREE through extensive experiments, comprehensive metrics and analysis \bc{on both synthetic and real-world datasets}.

\bc{Our main contributions are: (i) we introduce a new paradigm and formal formulation of lifelong learning under continual task drift for neural VRP solvers, motivated by observations of continual drift in real-world logistics data; (ii) we propose DREE, a general framework with dual replay and experience enhancement mechanisms for learning from this drift; and (iii) we validate DREE through extensive experiments on both synthetic continual-drift benchmarks and real-world datasets, supported by comprehensive assessment and analysis.}

\section{Related Work}

\subsection{Neural VRP Solvers}

By acquiring knowledge automatically from problem-solving experience, recent neural solvers lessen the dependence on human effort while achieving high problem-solving performance~\cite{ML4CO,HowGoodNCO,surveyL20,ML4VRP}.   
Two types of neural solvers exist, i.e., construction and improvement~\cite{AMDKD,N2S}.
Construction solvers (e.g.,~\cite{huang2025rethinking,POMO}) generate routes from scratch by choosing the next node step by step in an autoregressive process, and produce strong solutions within seconds. Improvement neural solvers instead begin with a complete solution and refine it through repeated updates, for example, by learning how to configure~\cite{WU2022learningImprovement,NeuOpt} or choose among predefined improvement operators~\cite{AOSsurvey,LearningAidedNS}. %LANSDOS
This work, we primarily study constructive neural solvers due to their broad applicability.

Existing neural solvers typically rely on sufficient one-off training with hundreds of epochs and millions of training instances sampled from a pre-defined, stationary task~\cite{POMO,INViT}. After training, model parameters are frozen and applied to solve unseen instances. A key challenge is generalization~\cite{cao2024towards,L2R}. While these solvers can perform well on unseen instances i.i.d. to training ones, their performance degrades sharply on out-of-distribution instances drawn from unseen tasks~\cite{HowGoodNCO,DROP,EL-DRL,Omni}. New model architectures (e.g., ~\cite{INViT,distanceawareattentioN,Lens,MTL-KD}) and learning strategies (e.g., ~\cite{ASP,liu2024prompt,TestTimeProjection}) are developed to improve generalization across tasks, yet they still rely on a representative pre-defined training task set and sufficient one-off learning on them, without actively reacting to potential arising new tasks.

\subsection{Lifelong Learning Neural VRP Solvers}

Fine-tuning is the standard approach when new tasks arise, namely training a solver pretrained on earlier tasks using instances from the new task~\cite{Omni,goal,CPLVRP,CoEKS}. However, fine-tuning leads to catastrophic forgetting, where parameters are overwritten to optimize the new task and performance drops substantially on previously learned tasks~\cite{surveyCL,surveyCRL}. To better balance and improve plasticity for acquiring new tasks and stability for retaining prior knowledge, recent work has begun to study lifelong learning for neural VRP solvers in settings where training tasks arise sequentially~\cite{li2024enhancing,feng2025lifelonglearner,LLR-BC}.

Buffering and replaying representative signals is an effective and widely adopted strategy to address drift~\cite{HallOfFame,SER,DER}.
Existing lifelong learning solvers~\cite{li2024enhancing,feng2025lifelonglearner,LLR-BC} typically take experience replay as a key mechanism to mitigate forgetting and facilitate knowledge transfer, but they differ in how experience is defined and replayed. \cite{li2024enhancing} and \cite{feng2025lifelonglearner} formulate problem instances as experience, assuming that the generation of new instances is controllable, and generate new instances of previous tasks to relearn old tasks during learning a new task. A sufficiently trained model for each task is also utilized to guide later learning on a new task. LLR-BC~\cite{LLR-BC} regards solver's behaviors as experience. By storing high-quality solution trajectories generated when learning previous tasks and encouraging the model to imitate these preserved behaviors during subsequent task learning, LLR-BC demonstrates strong plasticity, stability, and generalization when tasks vary significantly in both scale and distribution. 

Despite this progress, existing lifelong learning studies for neural VRP solver assume periodical stationary task. Their methods rely on sufficient per-task training, and obtaining high-quality task-specific models~\cite{li2024enhancing,feng2025lifelonglearner} or behaviors~\cite{LLR-BC} for subsequent knowledge consolidation. Issues arise in the continually drifting scenario. (i) A large number of tasks exist, each with highly limited training resources, making task-specific knowledge scarce and valuable. Using only instances or only behaviors as experience discards complementary information contained in the other. (ii) Without sufficient per-task training, the quality of buffered experience is substantially low. As a result, existing lifelong learning neural solvers are ineffective under the continually drifting scenario (cf. Section~\ref{sec:performance}).

\section{Lifelong Learning under Continually Drifting VRP Tasks}

\subsection{Continual Drift in VRP\label{sec:VPR_drift}}

\begin{figure}[h]
  \centering
   \vspace{-6pt} 
\includegraphics[width=0.5\linewidth]{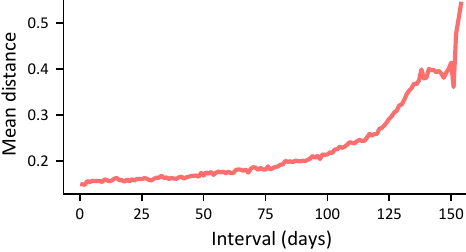}

   \vspace{-4pt} 
  \caption{Mean GW distance increases with the interval length, city Jilin, region 31. \label{fig:LaDe_GW_example}}
  % TODO
% \vspace{-10pt} 
\end{figure}

%\paragraph{Observed Real-world Drift}
\paragraph{Continual Drift Observed in Real World}

\bc{Based on LaDe~\cite{LaDe}, a real-world logistics dataset containing six months of daily delivery records from five cities with populations ranging from millions to tens of millions, we empirically verify the existence of continual drift in real-world applications. 
Specifically, for a region in a given city, which is typically served by one delivery station, we quantify the spatial discrepancy between customers on any two days using the Gromov--Wasserstein (GW) distance~\cite{GWdistance}. Then we draw the curve of the mean GW distance and the interval length between the two days to analyze the relation between them. 
% Although the GW distance curve exhibits slight fluctuations due to the inherent variability of real-world logistics data, it shows an overall increasing trend as the temporal interval between two days becomes larger (Figure~\ref{fig:LaDe_GW_example} shows an example from Jilin city).
With slight fluctuations, the GW distance curve shows an overall increasing trend as the temporal interval between two days becomes larger (Figure~\ref{fig:LaDe_GW_example} shows an example).
It reveals a clear continual drift trend in the spatial distribution of delivery customers.
Furthermore, the problem scale also shows continual drift trend. More details about the dataset, distance computation, and results are in Appendix~\ref{app:lade_benchmark}.}

\paragraph{Continual Drift in the View of Task}
\bc{Following the common formulation of VRP tasks in neural solver studies,} a VRP task $P$ in this work is defined as a problem instance distribution in fixed settings, such as scale~\cite{li2024enhancing,POMOMTL}, node-coordinate distribution~\cite{EL-DRL,LLR-BC}, constraints~\cite{zhou2024MVMoE,UniCO,ConstraintDegrees}, and the distance measurement~\cite{feng2025lifelonglearner}.
% Real-world application often exhibit continual task drift driven by factors such as business growth~\cite{MARKOV2020104798,math12010028,YTO}.
\bc{As observed, VRP instance distribution would continually drift.}
Under a fine-grained time discretization, the task can be approximately stationary within a single step, yet two consecutive steps have related but different tasks. 
% Reflecting practical cases, in the continually drifting lifelong learning scenario, the tasks arrived in different time step are different. 
In this case, learning of totally $T$ time steps is conducted on a sequence of tasks $\{P_{1},\dots,P_{T}\}$, where $P_{t}\neq P_{t'}$, $\forall t\neq t'$. Training at each step on the corresponding task is inherently insufficient, since neural solvers typically require a large number of steps (epochs) to learn a single stationary task sufficiently~\cite{li2024enhancing,feng2025lifelonglearner,POMO,INViT}. 

The periodically stationary scenario studied by existing works~\cite{li2024enhancing,feng2025lifelonglearner,LLR-BC} assumes each task remains unchanged for $n$ learning steps before the time step $t$ that the next task arrives, i.e., $P_{t’} = P_{t}$, $ \forall t'\in\{t-n,\dots,t-1\}$, where $n \gg 1$ so that sufficient per-task training is available.
Compared with it, the continually drifting scenario departs in that (i) no sufficient per-task training is guaranteed, and (ii) task distributions drift continuously over time rather than remaining constant over long intervals.

While plasticity is essential for adapting to new tasks, stability is also important. 
A solver should preserve performance on earlier tasks, as instances of them may re-occur and newly encountered tasks could be near-variants of past ones.
The overall goal is to learn a solver that can produce high-quality solutions for instances drawn from any seen task of any time step.

\subsection{Problem Formulation}

We formulate lifelong learning in the continually drifting scenario as lifelong reinforcement learning.
A constructive solver solve on a VRP instance $p$ by sequentially extending a partial solution, which can be modeled as a stationary instance-conditioned Markov decision process (MDP). Specifically, for an instance $p$, we define $\mathcal{M}(p)=\langle \mathcal{S}(p), \mathcal{A}(p), \mathcal{R}(p), \mathcal{T}(p)\rangle$,
where a state $s$ from the state space $S(p)$ encodes $p$ together with the current partial solution, an action $a$ from the action space $ \mathcal{A}(p)$ selects the next node to visit. The reward function $\mathcal{R}(p)$ is derived from the optimization objective (e.g., negative tour length~\cite{POMO}). $\mathcal{T}(p)$ specifies the state transition under the specific constraints.

A training instance $p$ is sampled from the task $P_t$ at time step $t$, i.e., $p\sim P_t$. We consider only node characteristics changing between instances of a task, leading to different state spaces but identical action space, reward function and transition dynamics for different instances, i.e., $\mathcal{S}(p)\neq \mathcal{S}(p')$, $\mathcal{A}(p) = \mathcal{A}(p')$, $\mathcal{R}(p) = \mathcal{R}(p')$,  $\mathcal{T}(p) = \mathcal{T}(p')$, $\forall p,p'\sim P_t$ and $p\neq p'$.
As an instance forms an MDP, a task $P_t$ introduces a distribution
$\mathbf{D}_t$ 
over the MDP, i.e., $\mathcal{M}(p) \sim \mathbf{D}_t$.

MDP distributions are different at different time steps, i.e., $\mathbf{D}_t\neq\mathbf{D}_{t'}$, $\forall t\neq t'$.
Changing the distance measure~\cite{feng2025lifelonglearner} or the objective functions~\cite{rethinkingNMOCO} changes the reward function, and changing constraints~\cite{CPLVRP,URS} (resulting in different problem variants) modifies the transition dynamics. 
This work focuses on drift in scale and node distribution, thus, distributions of $\mathcal{A}(p)$ and $\mathcal{S}(p)$ vary with $t$.

\section{DREE\label{sec:DREE}}

Experience replay of high-quality knowledge from previously seen tasks is a standard and effective strategy~\cite{li2024enhancing,feng2025lifelonglearner,LLR-BC} to mitigate catastrophic forgetting and promote knowledge transfer in lifelong learning scenarios. A good replay mechanism hinges on (i) what information is worth replaying as the experience, and (ii) how to obtain experience as high-quality as possible.

\begin{figure*}[t]
 \centering
 \includegraphics[width=.85\linewidth]{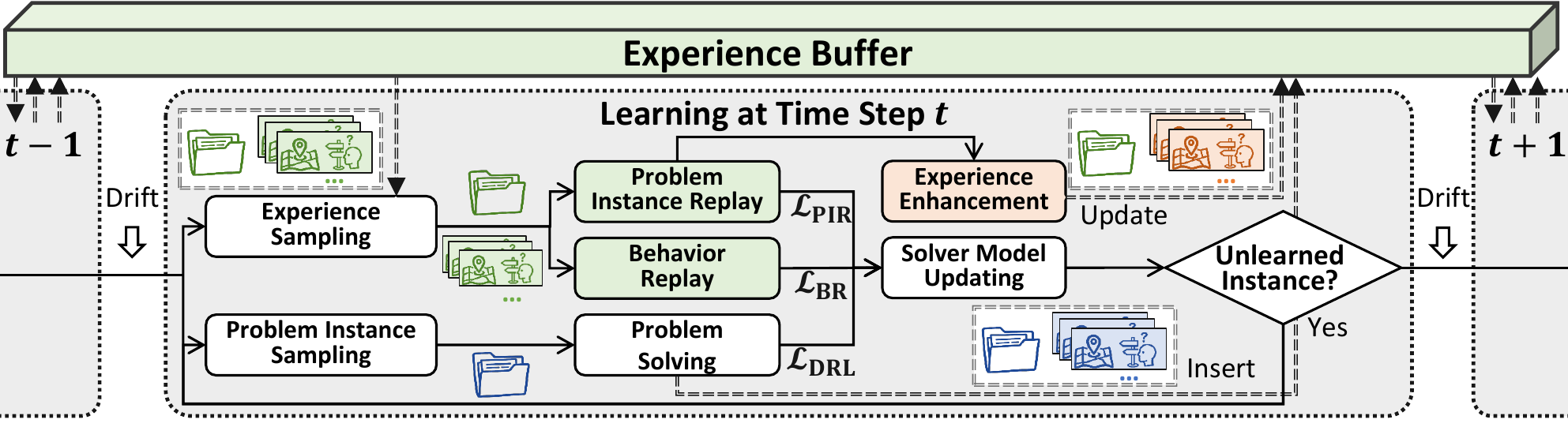}
 % \vspace{-12pt}
 \caption{Lifelong learning of DREE with the task drifts between every two consecutive time steps.}
 \vspace{-14pt}
 \label{fig:DREE}
\end{figure*}

To effectively and efficiently learn in the continually drifting scenario, 
we propose \textbf{DREE} (cf. Figure~\ref{fig:DREE}), a general lifelong learning framework with three key mechanisms: \textit{problem instance replay} (PIR), \textit{behavior replay} (BR), and \textit{experience enhancement} (EE).
Since per-task knowledge is scarce and valuable, DREE buffers both problem instances and the solver’s behaviors on these instances as experiences, to preserve rich knowledge.
During learning on a new task, DREE performs replay at both the instance level and behavior level to achieve effective knowledge consolidation. PIR enables the current solver to re-learn the buffered instances, reinforcing performance on past tasks. In parallel, BR regularizes learning by encouraging the solver to preserve high-quality behaviors on previous tasks, thereby transferring knowledge and reducing forgetting. 
To address the misleading of low-quality experiences, DREE actively improves the quality of buffered experiences. When the solver discovers a better solution for a buffered instance during PIR, the EE mechanism updates the stored behavior accordingly, continuously improving the quality of experiences. The following subsections introduce DREE's key components. Appendix~\ref{app:DREE} provides more details about DREE.

\subsection{Experience Buffer\label{sec:buffer}}

The base neural solver conducts the learning of an instance $p \sim P_t$ of a task $P_t$ to improve the performance on the task, by minimizing a loss $\mathcal{L}_{\mathrm{DRL}}(\theta,p)$,
where $\theta$ represents the solver model. The definition of $\mathcal{L}_{\mathrm{DRL}}(\theta,p)$ depends on the specific base solver used, and for different base solvers, the definition could be different. 
Experiences are generated while the solver learns the current task.

Each experience is defined as $e=\langle p, f, [(s,\pi)]\rangle$,
where $p$ denotes the problem instance, $f$ is the objective value of a solution of $p$, and $[(s,\pi)]$ represents the corresponding solution construction trajectory, i.e., a sequence of states $s$ and the solver’s action distributions (behaviors) $\pi$ on them. 

DREE maintains a buffer with a fixed size $|\mathcal{B}|$ and inserts newly generated experiences using reservoir sampling~\cite{reservoir}, so that each experience has an equal probability of being retained.
Specifically, a new experience is inserted directly if the buffer is not full. Otherwise, it replaces a random buffered experience with probability $\frac{|\mathcal{B}|}{|E|}$, where $|E|$ denotes the number of experiences generated so far.

As common practice, DREE trains with the problem instance in batch. Accordingly, experiences are also buffered and sampled at the batch level.
During learning a new task, DREE obtains new instance batches of the new task, and randomly samples experience from the buffer for PIR and BR after learning each new instance batch.
Instead of fixing the experiences once buffered~\cite{LLR-BC,SER,DER}, DREE actively updates them over time (cf. Section~\ref{sec:EE}) to pursue higher-quality knowledge.

\subsection{Problem Instance Replay\label{sec:PIR}}

A direct way to improve performance on previous tasks is to solve and learn instances of them again, analogous to training on new-task instances.
Prior works typically implement such replay by generating new instances of previous task during learning a new task~\cite{li2024enhancing,feng2025lifelonglearner}.
In contrast, DREE reuses the buffered instances, making it applicable when instance generation is uncontrollable.

Specifically, for a buffered instance $p_b$, DREE re-solves it using the current solver to obtain a new trajectory $[(s_\theta,\pi_\theta)]$ and a new solution with a new objective value $f(\theta,p_b)$.
The base solver then applies its standard training objective on this trajectory, yielding a problem instance replay (PIR) loss 
\[ 
\mathcal{L}_{\mathrm{PIR}}(\theta,e) = \mathcal{L}_{\mathrm{DRL}}(\theta,p_b),
\]
where $p_b$ is the buffered problem instance in $e$.
DREE performs PIR with an adaptive interval $N$, to balance plasticity and stability.
For a batch of $M$ replayed instances, let $M^+$ be the number of instances where the new objective is better than the buffered one.
DREE schedules the next PIR after learning $N \gets \; \mathrm{UB} - \frac{M^+}{M}\,(\mathrm{UB}-\mathrm{LB})$
batches of new-task instances, where $\mathrm{LB}$ and $\mathrm{UB}$ are hyperparameters.
Intuitively, a smaller $\frac{M^+}{M}$ indicates less potential to improve on previous tasks, triggering less frequent replay (larger $N$).
$\mathcal{L}_{\mathrm{PIR}}(\theta,e)\gets 0$ when PIR is not performed.

\subsection{Behavior Replay}

For constructive solvers, the solution and its quality are fully determined by the construction behaviors.
Thus, encouraging the solver to produce similar behaviors on the same (buffered) states is an effective and efficient way to preserve performance on previously learned tasks~\cite{LLR-BC,SER,DER}.
To achieve this, the behavior replay (BR) introduces an additional loss term, $\mathcal{L}_{\mathrm{BR}}(\theta, e)$, which measures the discrepancy between the buffered behaviors and those produced by the current solver. By minimizing $\mathcal{L}_{\mathrm{BR}}(\theta, e)$, DREE mitigates catastrophic forgetting when learning new tasks. 
Specifically, we conduct a problem instance-level behavior consolidation~\cite{LLR-BC} based on our experience definition, and define the behavior replay loss based on a buffered experience $e$ as 
\[ 
\mathcal{L}_{\mathrm{BR}}(\theta,e) = \sum_{s_b\in e} \bar{w}_{e}(s_b) \sum_{a\in \mathcal{A}} \pi_\theta(a\mid s_b)\, \log \frac{\pi_\theta(a\mid s_b)}{\pi_b(a\mid s_b)},
\]
where $\pi_\theta(\cdot\mid s_b)$ is the current solver’s action distribution at a buffered state $s_b \in e$, $\pi_b(\cdot\mid s_b)$ is the buffered behavior stored in $e$ corresponding to $s_b$, $\mathcal{A}$ is the corresponding action space, and $\bar{w}_{e}(s_b)$ is a confidence-based weight computed by CaEW~\cite{LLR-BC}.

\subsection{Experience Enhancement\label{sec:EE}}

Behavior replay is effective when the target behaviors are of high quality, i.e., they correspond to solutions with good objective values. In lifelong learning of VRP solvers, buffering behaviors only after sufficiently long training on a stationary task can ensure high-quality experiences~\cite{LLR-BC}.
However, in the continually drifting scenario, sufficient immediate per-task training is unavailable, leading to low-quality generated and buffered experiences. 
Since behavior replay encourages imitation of buffered behaviors, low-quality targets may misguide learning and hurt lifelong performance.

PIR provides an opportunity to improve behavior quality after experiences have been buffered.
It produces new solutions (and behaviors) whose objectives are directly comparable with the buffered ones.
We thus introduce an \textit{experience enhancement} mechanism.
After performing PIR, DREE compares the new objective $f(\theta,p)$ with the buffered objective $f_b$ for each replaced instance. Whenever $f(\theta,p)$ is better, i.e., $f(\theta,p)<f_b$, DREE updates the objective and the buffered state-behavior sequence with the newly generated ones, i.e., $f_b \gets f(\theta,p)$ and $[(s_b,\pi_b)] \gets [(s_\theta,\pi_\theta)]$.
As a result, BR always consolidates knowledge toward the best-so-far solutions and corresponding behaviors, preserving the best performance on past tasks.

Overall, DREE optimizes the loss:
\[
\mathcal{L}(\theta,p,e)
=
\mathcal{L}_{\mathrm{DRL}}(\theta,p)
+
\alpha\,\mathcal{L}_{\mathrm{BR}}(\theta,e)
+
\beta\,\mathcal{L}_{\mathrm{PIR}}(\theta,e),
\]
where $\theta$ is the current solver, $p \sim P_t$ is a training instance of the current task $P_t$, and $e$ is the experience sampled from the buffer.
$\alpha$ and $\beta$ are hyperparameters.

\section{Experiments\label{sec:experiment}}
Focusing on continual drift lifelong learning scenarios of neural VRP solvers, we conduct comprehensive experimental studies to answer the following research questions:
\begin{itemize}
\item \textbf{Effectiveness}: Can DREE effectively and efficiently solve instances from learned tasks?
\item \textbf{Stability and plasticity}: How does DREE perform in terms of stability and plasticity?
\item \textbf{Generalization}: Can DREE effectively solve problem instances from unseen tasks from different distributions and scales?
% \item \textbf{Impact of task changing type}: How does the difference in task changing impact the performance of existing methods and DREE?
\item \textbf{Hyperparameter sensitivity}: How sensitive is DREE to key hyperparameters: buffer size $|\mathcal{B}|$, the loss coefficients $\alpha$ and $\beta$, and the PIR interval range defined by $\mathrm{LB},\mathrm{UB}$?
\item \textbf{Applicability:} Can DREE also work effectively in other scenario (e.g., periodically stationary scenario), or, on different base neural solvers?
\end{itemize}

% We implement DREE\footnote{Code will be released upon acceptance of this paper.} majorly with POMO~\cite{POMO} as the base solver, given its concise design and broad applicability. Notably, DREE is a general framework and can be applied to other neural solvers. 

We implement DREE  majorly with POMO~\cite{POMO} as the base solver, given its concise design and broad applicability. Notably, DREE is a general framework and can be applied to other neural solvers.

\subsection{Experiment Setup\label{sec:experiment_setup}}

\paragraph{Datasets.}
% To our best knowledge, there is no available dataset tailored to the continually drifting lifelong learning scenario. We therefore construct a dataset based on the existing one designed for the periodically stationary scenario~\cite{LLR-BC}.
\bc{We conduct experiments on both synthetic and real-world datasets. The synthetic dataset is constructed based on an existing one of the periodically stationary scenario~\cite{LLR-BC}.
Specifically, for both TSP and CVRP, we consider five random orders over six principal tasks with different node distributions and scales.
We take an epoch (totally $T$ epochs) as the unit of time steps (which is small enough, cf. Appendix~\ref{app:synthetic_task}) and assign principal tasks evenly to the task sequence. The task of an epoch between two principal tasks is a linear mixture of them. 
Appendix~\ref{app:synthetic_dataset} provides more details about principal tasks and the mixture method.
In addition, TSPLIB~\cite{TSPLIB} and CVRPLIB~\cite{CVRPLIB}, are used for generalization ability evaluation. 
We further conduct an evaluation on a real-world dataset LaDe~\cite{LaDe}, taking each day as a time step, forming TSP instances with all customers served each day of a given region. We involve five regions to form five lifelong learning sequences, with one region for each city. The first 95\% days are used for lifelong learning, and the final 5\% are left for generalization evaluation. More details are in Appendix~\ref{app:lade_benchmark}. For both synthetic and real-world datasets, each task (one epoch) has 4,096 training instances, significantly less than the millions of instances used in existing studies for one task~\cite{POMO}, following the insufficient per-task training assumption.}

\paragraph{Training and Test Setting.}

% Hyperparameters are set as follows: $|\mathcal{B}|=256$ and $\beta=1$, where $|\mathcal{B}|$ are defined in units of batches.
% $\alpha=100$ for CVRP and $\alpha=10$ for TSP, as suggested by\cite{LLR-BC}.  
% % nips
% During training, an epoch is the unit of time step. All instances generated in one epoch are i.i.d.. 
% Training instances are sampled on the fly. We use a batch size of $M=32$, train 128 batches per epoch, and $T=1000$ epochs. 
% Thus, each stationary task has 4,096 training instances, which is significantly less than 
% a set of millions of instances used in existing studies for learning each task~\cite{POMO,LLR-BC}.
% Although the total number of training instances is the same, DREE involves additional problem-solving episodes due to PIR. For a fair comparison, we allow other lifelong learning methods to run more batches per epoch, ensuring that they perform the same number of episodes as DREE.
% For evaluation over different task orders, without loss of generality, we test the solver on the test instance set of each principal task, as principal tasks are consistent between orders, while intermediate tasks are not. Each principal task has 1000 test instances, which are consistent for all compared methods and task orders. 
% Additional details are provided in Appendix~\ref{app:synthetic_tt_setting}.

\bc{We set $|\mathcal{B}|=256$ and $\beta=1$.
$\alpha$ is set to 100 for CVRP and 10 for TSP, as suggested in \cite{LLR-BC}. The synthetic dataset has $T=1000$ epochs, while in the real-world dataset $T$ is determined by the valid days contained.
% In the real-world dataset, the instances are even fewer. 
% Although the number of training instances is the same, 
DREE involves additional problem-solving episodes due to PIR. For a fair comparison, we allow other lifelong learning methods to run more batches per epoch, ensuring that they perform the same number of episodes as DREE. For evaluation, without loss of generality, we test the solver on the test instance set of each principal task, which provides an efficient and effective evaluation (cf. Appendix~\ref{app:Fine-grained_test}). More settings are in Appendix~\ref{app:synthetic_tt_setting}.}

\paragraph{Comparison Methods.}
We include the commonly applied strategy in neural solver studies~\cite{Omni,CoEKS}, \textit{fine-tuning}, which sequentially trains the solver only on the current task at each time step. 
We then include existing lifelong learning solvers, which are all designed for the periodically stationary scenario, to the continually drifting scenario, including (ii) the method of~\cite{li2024enhancing} (denoted \textit{Li}) and (iii) LLR-BC~\cite{LLR-BC}. 
% We also evaluate (iv) a multi-task setting, \textit{MT-Ref}, which has access to all possible tasks and samples training instances by randomly switching $t$ in each batch~\cite{LLR-BC}. 
We also evaluate (iv) a multi-task setting, \textit{MT-Ref}, which has access to all possible tasks and randomly selects tasks to train in each batch. 
Notably, \textit{MT-Ref} only serves as an idealized reference, and it is not practically applicable in lifelong learning scenarios. 
We exclude other methods compared in~\cite{LLR-BC} (e.g., methods of ~\cite{feng2025lifelonglearner,LiBOG}), as they store one solver per task and repeatedly compare against all stored solvers. This is prohibitively expensive in continually drifting, where massive stationary tasks are encountered over time.
\bc{Appendix~\ref{app:baseline} presents more details about the compared methods and Appendix~\ref{app:further_baselie} provides additional experiment results of different settings of them.}

\begin{table*}[t]
\centering
\caption{Mean (std.) over five task orders. MT-Ref is not lifelong learning and is trained once, thus ABPl/AFB/AMBF and std. are not applicable.}
 % \vspace{-6pt}
\begin{adjustbox}{max width=\linewidth}
\begin{tabular}{c|c|c|c|c|c|c|c|c}
\toprule
{\multirow{2}{*}{Method}} & \multicolumn{4}{c|}{CVRP}& \multicolumn{4}{c}{TSP}\\
\cmidrule(lr){2-9}
   & AP & AFB   & AMFB   & ABPl &  AP& AFB   & AMFB   & ABPl \\
 \midrule
MT-Ref & \textbf{3.04} & - &- &- & \textbf{0.91}  &- &- &- \\
\midrule
Fine-tuning & 8.75 (4.40) & 5.47 (4.27) & 15.75 (6.13) & 3.28 (0.29) & 1.97 (0.30) & 0.86 (0.07) & 1.64 (0.32) & 1.11 (0.28) \\
 Li & 6.16 (0.38) & \textbf{0.09} \textbf{(0.06}) & \textbf{0.10} (\textbf{0.06}) & 6.07 (0.38) & 4.43 (0.96) & \textbf{0.02} (\textbf{0.01}) & \textbf{0.05} (\textbf{0.05}) & 4.42 (0.95)\\
 LLR-BC & 4.19 (0.32) & 1.10 (0.19) & 1.76 (0.18) & 3.09 (0.28) & 1.39 (0.29) & 0.34 (0.07) & 0.65 (0.20) & 1.05 (0.26)\\
% \midrule
\textbf{DREE}  & \textbf{3.11} (\textbf{0.13}) & 0.23 (0.15) & 0.42 (0.34) & \textbf{2.88} (\textbf{0.12}) & \textbf{1.19} (\textbf{0.27}) & 0.16 (0.05) &0.23 (\textbf{0.05})& \textbf{1.03} (\textbf{0.25})\\
\bottomrule
\end{tabular}
\end{adjustbox}
 \vspace{-12pt}
\label{tab:measures}
\end{table*}

\paragraph{Evaluation Metrics.}

We evaluate a lifelong learning solver with a comprehensive set of metrics, adapted from metrics designed for the periodically stationary scenario~\cite{LLR-BC}.
Let $d_{t,i}$ denote the \textit{test performance}, i.e., average optimality gap (\%) over all test instances, of the solver on principal task $P_i$ after $t$th training epoch. 
${d}^{*}_{i}$ denotes the best (smallest) test performance on a principal task $P^i$ in one lifelong learning process, and $t^{*}_{i}$ denotes the epoch that ${d}^{*}_{i}$ is achieved (i.e., ${d}^{*}_{i} = {d}_{t^{*}_i,i}$).
Based on this notation, after all $T$ epochs of one lifelong learning, we calculate (\textbf{smaller is better for all}):
\begin{itemize}
% \vspace{-6pt}
 \item \textbf{Average Performance (AP)}: the average performance on tasks learned, i.e., $AP=\frac{1}{K}\sum_{i=1}^{K} {d}_{E,i}$.
  \item \textbf{Average Forgetting after Best (AFB)}: the performance decrease on previously learned tasks after the best performance of this task is achieved, i.e., $AFB=\frac{1}{K}\sum_{i=1}^{K} {d}_{E,i}-{d}^{*}_{i}$.
  \item \textbf{Average Max Forgetting after Best (AMFB)}: the maximal forgetting of each learned task during the lifelong learning, i.e., $AMFB=\frac{1}{K}\sum_{i=1}^{K} \max_{t=e^{*}_{i}}^{E }{d}_{t,i}-{d}^{*}_{i}$.
   \item \textbf{Average Best Plasticity (ABPl)}: the average of best performance a solver achieves on each task, i.e., $ABPl=\frac{1}{K} \sum_{i=1}^K {d}^{*}_{i}$.
 % \vspace{-6pt}
\end{itemize}

\subsection{Performance on Synthetic Continual Drift\label{sec:performance}}

\paragraph{Performances on Seen Tasks.}

Table~\ref{tab:measures} demonstrates the metrics of each method. In terms of AP, DREE stably (smallest std.) outperforms (smaller mean) all compared lifelong learning methods over task orders on both TSP and CVRP. 
DREE even performs closely to \textit{MT-Ref}, which can access all tasks simultaneously. 
It verifies DREE's effectiveness in solving tasks after lifelong learning of them.

% \paragraph{Plasticity during Learning.}

% DREE attains the smallest ABPl on CVRP and the second smallest on TSP, indicating strong plasticity. This suggests that the BC, IR, and EE mechanisms effectively transfer knowledge from previous tasks to facilitate learning on new tasks.

% \paragraph{Stability during Learning.}
% \textit{Fine-tuning} exhibits severe forgetting (large AFB and AMFB), as expected. \textit{Li} strongly biases toward stability, yielding very small forgetting but substantially sacrificing plasticity (large AP and ABPl).
% In contrast, while maintaining promising plasticity, DREE achieves lower forgetting than all compared methods except \textit{Li} for both TSP and CVRP in terms of AFB and AMFB, highlighting the effectiveness of the key modules of DREE (BC, IR, and EE) in mitigating forgetting.
% % (see also Figure~\ref{fig:forgetting}).
% We record the solver after each epoch, and Figure~\ref{fig:forgetting} shows their test performance. DREE effectively improves performance on new tasks while stably preserving, sometimes even improving, its performance on previously learned tasks.

\begin{figure*}[t]
\centering
\begin{subfigure}{.46\linewidth}
 \centering
 \includegraphics[width=\linewidth]{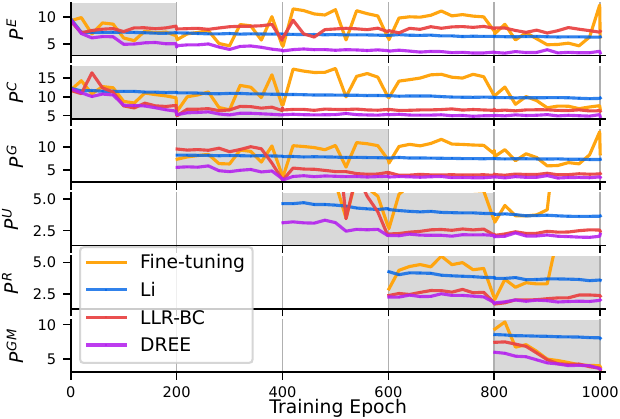}
 \vspace{-10pt}
 \caption{CVRP}
\end{subfigure}
\hspace{18pt}
\begin{subfigure}{.45\linewidth}
 \centering
 \includegraphics[width=\linewidth]{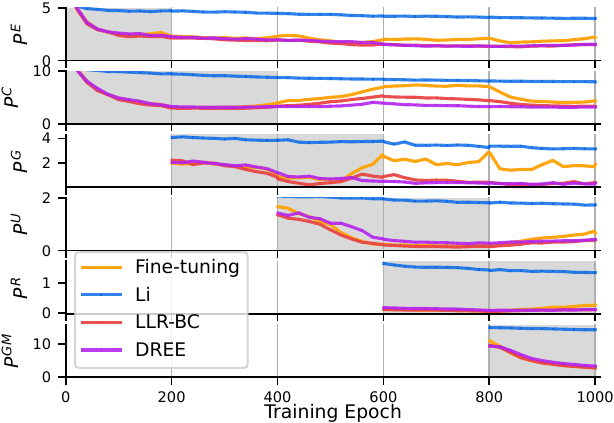}
 \vspace{-10pt}
 \caption{TSP}
\end{subfigure}

 \vspace{-6pt}
\caption{Learning curve of task order 1, measured by test performance. Grey background indicates the epochs that the corresponding principal task is involved in generating intermediate tasks. }
% todo
 \vspace{-10pt}
\label{fig:forgetting}
\end{figure*}

\paragraph{Plasticity and Stability.}
% todo
DREE attains overall the smallest ABPl, indicating its strong plasticity. This suggests that the BR, IR, and EE mechanisms effectively transfer knowledge from previous tasks to facilitate learning on new tasks. \textit{Fine-tuning} exhibits severe forgetting (large AFB and AMFB), as expected. \textit{Li} strongly biases toward stability, yielding very small forgetting but substantially sacrificing plasticity (large AP and ABPl).
In contrast, while maintaining promising plasticity, DREE achieves lower forgetting than all compared methods except \textit{Li}, highlighting the effectiveness of the key modules of DREE (BR, IR, and EE) in mitigating forgetting.
Figure~\ref{fig:forgetting} shows test performance curve during lifelong learning. DREE effectively improves performance on new tasks while stably preserving its performance on previously learned tasks.

% \begin{table}[htbp]
\begin{wraptable}{r}{0.4\textwidth}
 \vspace{-10pt}
\centering
\caption{Mean (Std.) gap on unseen benchmarks. }
% todo
\begingroup
% \footnotesize   
\begin{adjustbox}{max width=\linewidth}
\begin{tabular}{c|c|c}
\toprule
{\multirow{1}{*}{Method}} & CVRPLIB&TSPLIB\\
\midrule
MT-Ref & \textbf{12.92 }(\textbf{4.90}) & 19.86 (10.69) \\
\midrule
Fine-tuning  & 16.32 (\textbf{6.39}) & 19.74 (9.56) \\
 Li & 21.66 (7.13) & 30.20 (13.76) \\
LLR-BC  & 35.57 (31.83) & 18.96 (9.59)\\ %& 18.91 (9.60) \\
 % \midrule
\textbf{DREE}  & \textbf{13.78}  (6.40) & \textbf{18.06} (\textbf{9.40}) \\%\textbf{18.77} (\textbf{9.46})\\
\bottomrule
\end{tabular}
\end{adjustbox}
\endgroup
 \vspace{-16pt}
\label{tab:libs}
\end{wraptable}
% \end{table}

\paragraph{Performances on Unseen Benchmark Instances.}

Table~\ref{tab:libs} lists the test performance on CVRPLIB and TSPLIB of solvers lifelong learned from task order 1. DREE outperforms (smaller gap) all compared lifelong learning methods
and obtains close (TSP) or even better (CVRP) performance than \textit{MT-Ref},
verifying its promising generalization ability by capturing general knowledge across tasks. Appendix~\ref{app:libs} presents more details.

\subsection{Applicability~\label{sec:applibality}}

\paragraph{To Different Base Solvers.}

% \begin{table}[htbp]
\begin{wraptable}{r}{0.6\textwidth}
\centering
 \vspace{-8pt}
\caption{Metric values with Omni or INViT as base solver. 
}
 % \vspace{-8pt}
\label{tab:applicability}
\begin{adjustbox}{max width=\linewidth}
 \setlength{\tabcolsep}{3pt}
\begin{tabular}{c|c|c|c|c|c|c|c|c}
\toprule
{\multirow{2}{*}{Method}} & \multicolumn{4}{c|}{CVRP}& \multicolumn{4}{c}{TSP}\\
\cmidrule(lr){2-9}
   & AP & AFB   & AMFB   & ABPl &  AP& AFB   & AMFB   & ABPl \\
\midrule
 % Fine-tuning (INViT) & 24.65 & 0.67 & 0.84 & 23.98 & 12.65 & 0.25 & 0.41 & 12.40\\
 % LLR-BC (INViT) & 23.83 & 0.33 & 0.66 & 23.49 & 11.78 & 0.25 & 0.62 & 11.54\\
 % \textbf{DREE (INViT)} & \textbf{21.93} & \textbf{0.24} & \textbf{0.56} & \textbf{21.69} & \textbf{10.99} & \textbf{0.16} & \textbf{0.21} & \textbf{10.83} \\
  Fine-tuning (INViT) & 23.01 & 1.00 & 1.46 & 22.02 & 9.00 & 0.47 & 0.63 & 8.52\\
 LLR-BC (INViT) & 21.87 & 0.30 & 0.45 & 21.57 & 9.97 & 0.39 & 0.74 & 9.57\\
 \textbf{DREE (INViT)} & \textbf{21.06} & \textbf{0.18} & \textbf{0.36} & \textbf{20.88} & \textbf{8.45} & \textbf{0.32} & \textbf{0.56} & \textbf{8.13} \\
\midrule
 Fine-tuning (Omni) & 5.67 & 2.46 & 7.63 & 3.20 & 1.02 & 0.31 & 1.04 & \textbf{0.71}\\
 LLR-BC (Omni) & 3.98 & 1.07 & 2.52 & \textbf{2.91} & 1.08 & 0.07 & 0.09 & 1.01\\
 \textbf{DREE (Omni)} & \textbf{3.45} & \textbf{0.54} & \textbf{1.52} & \textbf{2.91} & \textbf{0.98} & \textbf{0.06} & \textbf{0.07} & 0.93\\
\bottomrule
\end{tabular}
\end{adjustbox}
\end{wraptable}

% DREE is a general framework and can be applied to different neural solvers. 
To verify that the superior performance of DREE over other lifelong learning solvers does not stem from a bias toward a specific base solver (e.g., POMO), we conduct a comparison based on two other existing solvers (i.e., Omni~\cite{Omni} and INViT~\cite{INViT}). We implement and compare DREE with \textit{fine-tuning} and LLR-BC on task order 1.  Our goal is not to compare across base solvers but to evaluate different lifelong learning methods on the same base solver. The results across different base solvers are not directly comparable, as training settings are different (cf. Appendix~\ref{app:applicability_results}).
As Table~\ref{tab:applicability} shows, DREE outperforms \textit{fine-tuning} and LLR-BC generally, with a similar pattern found on POMO.
In summary, DREE is effective in the continually drifting scenario, without relying on specific characteristics of the base solver.

 \paragraph{\bc{To Scenarios with Additional Task Change Types.}}

\bc{We further conduct comparison on (i) the periodically stationary scenario~\cite{LLR-BC}, and (ii) a variant of the continual drifting scenario with sudden changes, in which we randomly select three epoch intervals lengths and swap their corresponding tasks, thereby introducing multiple abrupt changes into the task sequence (more details in Appendix~\ref{app:sudden_change}).
Table~\ref{tab:measures_more_scenarios} reports the results, based on task order 1. In both scenarios, DREE outperforms compared methods, with a similar pattern found in the continually drifting scenario.  This demonstrates the robustness of DREE to more complex drift dynamics.}

\begin{table}[htbp]
% \begin{wraptable}{r}{0.51\textwidth}
\centering
 % \vspace{-20pt}
\caption{Results on two additional scenarios. }
\begin{adjustbox}{max width=\linewidth}
 \setlength{\tabcolsep}{2.5pt}
\begin{tabular}{c|c|c|c|c|c|c|c|c|c|c|c|c|c|c|c|c}
\toprule
{\multirow{4}{*}{Method}} & \multicolumn{8}{c|}{Periodically   Stationary} & \multicolumn{8}{c}{Continual Drifting   with Sudden Changes}   \\
\cmidrule(lr){2-17}
& \multicolumn{4}{c|}{CVRP} & \multicolumn{4}{c|}{TSP} & \multicolumn{4}{c|}{CVRP} & \multicolumn{4}{c}{TSP}  \\
   
\cmidrule(lr){2-17}
 & AP& AFB& AMFB   & ABPl   & AP& AFB   & AMFB   & ABPl   & AP & AFB   & AMFB   & ABPl   & AP & AFB& AMFB   & ABPl  \\
 \midrule
Fine-tuning & 6.67  & 4.34   & 6.44   & 2.34   & 2.40   & 1.68  & 4.13   & 0.73   & 6.52   & 3.36  & 8.67   & 3.16   & 1.55   & 0.81   & 0.93   & 0.74  \\
LLR-BC  & 2.55  & 0.28   & 0.31   & 2.27   & 1.46  & 0.79  & 1.47   & \textbf{0.67} & 4.31   & 1.14  & 2.76   & 3.18   & 0.90& \textbf{0.11} & 0.13   & 0.79  \\
\textbf{DREE}  & \textbf{2.30} & \textbf{0.04} & \textbf{0.09} & \textbf{2.26} & \textbf{0.80} & \textbf{0.10} & \textbf{0.16} & 0.70& \textbf{3.19} & \textbf{0.30} & \textbf{0.51} & \textbf{2.89} & \textbf{0.81} & \textbf{0.11} & \textbf{0.11} & \textbf{0.70} \\
\bottomrule
\end{tabular}
\end{adjustbox}
 \vspace{-8pt}
\label{tab:measures_more_scenarios}
\end{table}

\subsection{Ablation Study and Hyperparameter Analysis\label{sec:ablation}}

% \paragraph{Comparison with Ablation Versions.}

% We form ablation versions by removing PIR, BR, and EE individually, denoted as \textit{w.o.PIR}, \textit{w.o.BR}, and \textit{w.o.EE}, respectively. More details are in Appendix~\ref{app:ablation}.
% We run them on task order 1, and Table~\ref{tab:parameter} lists the results.
% Removing each key component leads to performance degradation across multiple metrics on both CVRP and TSP, confirming the contribution of each module. 

% \paragraph{Comparison between Different Hyperparameter Settings.}

% We vary the values of key hyperparameters and evaluate DREE under each setting, as reported in Table~\ref{tab:ablation}. Additional details are provided in Appendix~\ref{app:hyperparameter}. We denote a variant with hyperparameter $k$ set to $v$ as $k(v)$.
% Overall, varying these hyperparameters causes only minor fluctuations across metrics, considering the performance gap between DREE and the baselines. This suggests that DREE is not overly sensitive to its hyperparameter settings.

\begin{wraptable}{r}{0.5\textwidth}
\centering
\vspace{-8pt}
\caption{Results of ablation study. 
\label{tab:ablation}
}
\vspace{-4pt}
\begin{adjustbox}{max width=\linewidth}
 \setlength{\tabcolsep}{3.5pt}
\begin{tabular}{c|c|c|c|c|c|c|c|c}
\toprule
\multirow{2}{*}{Method}& \multicolumn{4}{c|}{CVRP}& \multicolumn{4}{c}{TSP} \\ \cmidrule(lr){2-9}
& AP  & AFB  & AMFB & ABPl  & AP  & AFB  & AMFB & ABPl   
\\ \midrule

w.o.PIR  & 4.07 & 0.81 & 2.23 & 3.25 & 1.38 & 0.28 & 0.42 & 1.10 \\
w.o.BR & 3.87 & 0.53 & 1.70 & 3.33  & 1.35 & 0.29 & 0.48 & 1.06\\
w.o.EE  & 3.47 & 0.32 & 0.49 & 3.16 & 1.20 & 0.17 & 0.42 & 1.03\\
 \midrule
 \textbf{DREE} & \textbf{3.16} & \textbf{0.22} & \textbf{0.42} & \textbf{2.94}  & \textbf{1.12} & \textbf{0.10} & \textbf{0.29} & \textbf{1.02}\\
\bottomrule

\end{tabular}
\end{adjustbox}
 \vspace{-8pt}
\end{wraptable}

\bc{We form ablation versions by removing PIR, BR, and EE individually (cf. Appendix~\ref{app:ablation}).
We run them on task order 1, and Table~\ref{tab:ablation} lists the results.
Removing each component leads to performance degradation across multiple metrics, confirming the contribution of them. We further evaluate DREE with different hyperparameter settings, with results reported in Appendix~\ref{app:hyperparameter}. Overall, varying these hyperparameters causes only minor fluctuations across metrics, suggesting that DREE is not overly sensitive to its hyperparameter settings.}

\begin{wraptable}{r}{0.6\textwidth}
 \vspace{-10pt}
\centering
\caption{Mean (std.) over the real-world dataset.}
 \vspace{-4pt}
\begin{adjustbox}{max width=\linewidth}
\begin{tabular}{c|c|c|c|c|c}
\toprule
% Method & AP & AFB& AMFB& ABPl & Gen \\
%  \midrule
% MT-Ref & 1.93 (1.53) & - &- &- & 1.48 (\textbf{0.36})  \\
% \midrule
% Fine-tuning & 5.55 (4.60)&4.67 (4.36)&10.31 (7.26)&0.88 (0.58)&2.08 (1.49) \\
%  Li & 5.49 (3.71)&1.97 (2.75)&4.80 (5.62)&3.53 (1.45)&7.00 (2.64)\\ 
%  LLR-BC&1.85 (1.25)&1.49 (1.02)&3.51 (2.25)&0.36 (\textbf{0.24})&1.81 (0.99)\\
% % \midrule
% \textbf{DREE}&  \textbf{1.12} (\textbf{0.77})&	\textbf{0.93} (\textbf{0.58})& \textbf{2.32} (\textbf{1.69})& \textbf{0.19} (\textbf{0.24})& \textbf{1.24} (0.51) \\
Method  & AP & AFB& AMFB  & ABPl  & Gen\\ \midrule
MT-Ref  & 1.06 \textbf{(0.48)} & -  & -  & -  & 1.47 \textbf{(0.47)} \\ \midrule
Fine-tuning& 4.64 (3.79) & 3.82 (3.18) & 12.90 (13.88)  & 0.82 (0.65) & 3.16 (2.66) \\
Li& 10.02 (5.23)& 4.46 (5.25) & 8.32 (8.26) & 5.56 (1.81) & 12.67 (8.98)\\
LLR-BC  & 1.35 (1.29) & 1.03 (1.04) & 2.59 (1.97) & 0.32 (0.26) & 1.57 (1.19) \\
\textbf{DREE} & \textbf{1.03 (0.80)} & \textbf{0.73 (0.58)} & \textbf{2.06 (1.88)} & \textbf{0.31 (0.23)} & \textbf{1.36 (0.66)} \\
\bottomrule
\end{tabular}
\end{adjustbox}
 \vspace{-6pt}
\label{tab:LaDe_results}
\end{wraptable}

\subsection{\bc{Evaluation on Real-World Dataset} \label{sec:LaDe_performance}}
\bc{We further conduct lifelong learning on the real-world dataset and Table~\ref{tab:LaDe_results} lists the overall results (more details in Appendix~\ref{app:LaDe_result_details}). DREE still consistently outperforms the compared methods. In particular, DREE preserves the strong stability and plasticity, and also achieves the best performance (denoted \textit{Gen}, smaller better) to unseen instances. Moreover, by reinforcing high-quality learned behaviors, DREE achieves better performance than MT-Ref. It demonstrates DREE's effectiveness in handling drift in real-world logistics scenarios.}

\section{Conclusions\label{sec:conclusion}}
\bc{As observed in the real-world logistics dataset, new VRP instances commonly sequentially arrive with the problem pattern/distribution continually drifting.}
We propose lifelong learning for neural VRP solvers under this continual task drift, and formulate it as a lifelong RL problem with a drifting MDP distribution. 
\bc{A major challenge in this case is the low experience quality due to the frequent task change and insufficient per-task learning.}
To address it, we propose DREE, which integrates problem instance replay, behavior replay, and experience enhancement to continually acquire, rehearse, and refine experience.
Experiments \bc{on synthetic and real-world datasets} verify that DREE can achieve superior performance to existing lifelong learning solvers under the continually drifting scenario, \bc{while remaining effective on more types of task drifting.}
This highlights that, in lifelong learning of VRP solvers, both problem instances and behaviors constitute critical experience, and enhancing the quality of buffered experience is crucial for effective learning.

Despite its strong results, DREE treats tasks of all time steps equally. A potential future direction is to selectively emphasize steps that accumulate significant drift and are most distinctive to others for better efficiency. Incorporating strategies from dynamic optimization~\cite{10.1145/2739482.2756589} or online learning~\cite{HierarchicalDrfit} also has potential for improvement. Extending DREE to scenarios with drifting constraints, so as to lifelong learn diverse problem variants, also constitutes a promising future direction.

% TODO: check all reference, update arxiv?
\bibliographystyle{unsrtnat}
\bibliography{main}

%%%%%%%%%%%%%%%%%%%%%%%%%%%%%%%%%%%%%%%%%%%%%%%%%%%%%%%%%%%%
\clearpage
\appendix

\section{Details of DREE~\label{app:DREE}}

\subsection{General Process}
Algorithm~\ref{algo:DREE} demonstrates the process of DREE.
For each learning time step (epoch), following common practice~\cite{POMO,Omni,LLR-BC}, DREE iteratively learns in units of batch. In each iteration, a batch of new instances of the current task is sampled, obtaining $\mathcal{L}_{\mathrm{DRL}}$ to learn the current task. One experience batch is sampled, on which $\mathcal{L}_{\mathrm{BR}}$ is obtained to learn from buffered behavior.  PIR is conditionally conducted based on $N$, to better balance between plasticity and stability.
To calculate each loss term (i.e., $\mathcal{L}_{\mathrm{DRL}}$, $\mathcal{L}_{\mathrm{BR}}$, or $\mathcal{L}_{\mathrm{PIR}}$) based on an instance/experience batch, DREE takes the average value over the batch.
\begin{algorithm}[h]
\caption{DREE Lifelong Learning~\label{algo:DREE}}
\begin{algorithmic}[1]
\STATE \textbf{Input:} $\{P_t\}_{t=1}^T,\theta_0$
% \STATE \textbf{Parameter:} $N, N_D,M,\lambda,\beta$
\STATE \textbf{Parameters: }$\alpha,\beta,\mathrm{UB},\mathrm{LB},|\mathcal{B}|$
\STATE \textbf{Output:} $\theta$
\STATE Initialize buffer $\mathcal{B}\gets \emptyset$
\FOR{$t \in \{1, \dots, T\}$}
\FOR{$i \in \{1, \dots, I\}$}
\STATE $\{p\} \gets$ a batch of $M$ instances from $P_t$
\STATE $\{[s_\theta,\pi_\theta], f(\theta,p)\} \gets$ Solve $\{p\}$ with ${\theta}$
\STATE $\mathcal{L}_{\mathrm{DRL}} \gets $ base solver loss
\IF{$\mathcal{B}$ is not empty}
  \STATE $\{e\} \gets$ uniformly sample 
  \STATE $\mathcal{L}_{\mathrm{BR}}\gets $ BR
  \IF{$ i - i' = N$}
\STATE $\mathcal{L}_{\mathrm{PIR}}, M^+ \gets $ PIR
\STATE $i' \gets i$
\STATE $N \gets \; \mathrm{UB} - \frac{M^+}{M}\,(\mathrm{UB}-\mathrm{LB})$
  \ELSE
\STATE $\mathcal{L}_{\mathrm{PIR}} \gets 0$
  \ENDIF
\ELSE
  \STATE $\mathcal{L}_{\mathrm{PIR}} \gets 0$, $\mathcal{L}_{\mathrm{BR}} \gets 0$
\ENDIF
\STATE $\mathcal{L} \gets \mathcal{L}_{\mathrm{DRL}}+\alpha \mathcal{L}_{\mathrm{BR}}+ \beta \mathcal{L}_{\mathrm{PIR}}$
\STATE $\theta \gets$ Optimize($\theta$, $\mathcal{L}$)
\STATE $\mathcal{B} \gets $ reservoir sampling $(\{\{[s_\theta,\pi_\theta]\} \},\mathcal{B})$
\STATE update $e$ with $\{\langle p, f(\theta,p), [(s_\theta,\pi_\theta)]\rangle\}$ for any $f(\theta,p) < f_b$
\ENDFOR
\ENDFOR
\STATE \textbf{return} $\theta$
\end{algorithmic}
\end{algorithm}

\subsection{Hyperparameters}

DREE introduces five new hyperparameters, i.e., the coefficients of PIR and BR loss terms $\alpha$ and $\beta$, the buffer size $|\mathcal{B}|$, and the upper and lower bound of PIR interval $\mathrm{UB}$ and $\mathrm{LB}$. We set $\alpha = 100$ for CVRP and $\alpha = 10$ for TSP, which have the best performance in~\cite{LLR-BC}. Intuitively, we set $\beta = 1$ as PIR is analogous to training on new-task instance (obtaining $\mathcal{L}_{\mathrm{DRL}}$). 

The $\mathrm{UB}$ and $\mathrm{LB}$ determine the admissible range of $N$, thereby affecting the overall computational overhead of PIR throughout lifelong learning as well as the resulting plasticity–stability trade-off. In our implementation, we set this range heuristically to 1–4. As demonstrated in our experiments (cf. Section~\ref{sec:ablation}), DREE can automatically adapt its behavior, making its performance relatively insensitive to this choice.

$|\mathcal{B}|$ controls the buffer’s coverage of instances from previous tasks as well as the diversity of stored instances. Intuitively, a larger $|\mathcal{B}|$ can improve performance, but it also incurs higher storage cost. We set $|\mathcal{B}| = 256$ because it provides broad coverage while maintaining acceptable storage overhead.

\subsection{Computational Overhead \label{app:overhead}}

The proposed novel PIR and EE modules have promising efficiency, enabling DREE to outperform the SOTA baseline LLR-BC with less training time and fewer batches.
Taking task order 1 of the synthetic CVRP dataset as an example, DREE takes 151.26  wall-clock seconds per epoch on average, of which PIR accounts for 26.77\% of the total training time, while the overhead of EE is negligible since it only involves a few scalar comparisons. For comparison, LLR-BC takes 170.53 seconds per epoch. DREE takes less time because it uses fewer batches per epoch for fair comparison. More detailed wall-clock time records are listed in Table~\ref{tab:wall-clock}.

\begin{table}[htbp]
\centering
\caption{Wall-clock time (seconds) of PIR and its corresponding percentage to the wall-clock time of fine-tuning, under different task orders. Where GD refers to the gradually drifting scenario, and PS refers to the periodically stationary scenario. }
\begin{tabular}{c|c|c|c|c|c|c|c}
\toprule
\multirow{2}{*}{Base } & \multirow{2}{*}{Task }& \multicolumn{3}{c|}{CVRP}& \multicolumn{3}{c}{TSP} \\ \cmidrule(lr){3-8}
Solver & Order & PIR & Fine-tuning &  Percentage & PIR & Fine-tuning &  Percentage \\
  \midrule
\multirow{7}{*}{POMO} & GD1 & 40.49 & 88.81 & 45.59\%  & 28.99 & 58.96 & 49.17\% \\
 &GD2 & 30.91 & 73.63 & 41.98\% & 31.27 & 61.00 & 51.27\%  \\
 &GD3 & 45.65 & 89.01 & 51.29\%  & 18.33 & 58.27 & 31.45\% \\
 &GD4 & 40.80 & 76.55 & 53.29\%  & 25.34 & 51.33 & 49.37\% \\
 &GD5 & 47.84 & 88.69 & 53.94\%  & 27.77 & 58.06 & 47.83\% \\
 &PS1 & 20.02 & 73.65 & 27.18\%  & 19.56 & 59.90 & 33.15\% \\
 \cmidrule(lr){2-8}
 & Avg. & 37.61 & 81.72 & 46.02\% & 25.21 & 57.92 & 43.52\% \\ 
 \midrule
 INViT & GD1 & 53.26 & 139.35 & 38.11\%  & 41.87 & 95.04 & 44.05\% \\
\bottomrule
\end{tabular}
\label{tab:wall-clock}
\end{table}

The computational time cost of DREE majorly depends on the architecture of the base neural solver. 
Once PIR has the exact same complexity as the base solver to learn on a batch of new instances, while the complexity of BR varies.
In terms of FLOPs, using POMO as the base solver, once BR takes about 2.85 billion FLOPs for scale 20 CVRP experiences with batch size 32 and three depot returns. With $N$ as the problem scale of the experience and  $m$ as the number of depot returns, the FLOPs of once BR with different base solvers are listed in Table~\ref{tab:FLOP}

\begin{table}[htbp]
\centering
\caption{Computational cost of BR under different cases and base solvers, in terms of FLOPs. Notably, Omni and POMO share the same model architecture, thus the same cost.}
\begin{tabular}{c|c|c|c|c}
\toprule
  Base solver & Problem & $N=20$ & $N=50$ & $N=100$ \\
  \midrule
  POMO & CVRP & 2.84B + 4.26M $\times \ m$ & 7.20B + 5.46M $\times \ m$ & 15.28B + 7.45M $\times \ m$  \\
POMO & TSP & 2.73B & 7.14B & 15.28B  \\
INViT & CVRP & 78.53B+4.13B$\times \ m$ & 300.79B+6.14B$\times \ m$ & 607.71B+6.14B$\times \ m$  \\
INViT & TSP & 78.53B & 300.79B & 607.71B  \\
\bottomrule
\end{tabular}
\label{tab:FLOP}
\end{table}

Notably, the cost of BR, PIR, and EE does not grow with buffer size, since only one item is sampled from the buffer at a time.

During testing/deployment, since all compared lifelong learning methods with the same base neural solver share the same model architecture, the difference in inference cost is negligible.

\subsection{Novelty and Difference to Existing Lifelong Learning Solvers}
DREE has significant differences from the method of \cite{li2024enhancing} and \cite{feng2025lifelonglearner} as it does not actively generate new instances, and it does not require storing any learned model parameters. Instead, DREE buffers instance and behavior for knowledge consolidation.

Although DREE adapts the effective behavior consolidation mechanism of LLR-BC for behavior replay, it is substantially novel as it integrates PIR and BR to achieve more comprehensive and effective knowledge consolidation, while further introducing EE to address a key challenge in continually drifting scenarios: the accumulation of low-quality experiences. Therefore, rather than simply extending LLR-BC, DREE develops a dedicated framework for continual task drift, with several important differences from LLR-BC, as detailed below.
\begin{itemize}
\item DREE defines experience at the instance level, represented by an instance, its objective value, and the corresponding solution trajectory, thereby retaining richer information. In contrast, LLR-BC defines experience at the step level, using only state–behavior pairs.
\item DREE replays problem instances to enable further exploration for better solutions, whereas LLR-BC does not support this type of replay.
\item By comparing objective values, DREE can assess and compare the quality of stored experiences, which facilitates continuous improvement of experience quality. LLR-BC does not provide such a mechanism.
\item In DREE, buffered experiences can be updated when better solutions are found. In LLR-BC, however, experiences remain fixed once stored, regardless of their quality.
\end{itemize}

\section{Compared methods and implementation details.\label{app:baseline}}
All compared methods are implemented on top of the same base neural solver to ensure a fair comparison. All methods are trained from the same randomly initialized model. We use instance augmentation~\cite{POMO} with a factor of 8 during learning. All experiments are conducted on a GPU cluster utilizing a single Nvidia A100 with 512 GB memory. A full synthetic run took approximately 2 days, and the real-world run took approximately 12 GPU-hours. 

\paragraph{LLR-BC.} LLR-BC \cite{LLR-BC} is a behavior-level lifelong learning method for neural VRP solvers, designed to mitigate catastrophic forgetting while preserving the solver's ability to adapt to newly arriving tasks. Instead of constraining model parameters directly, LLR-BC stores past solving experiences in a fixed-size buffer maintained by reservoir sampling and consolidates previous knowledge by aligning the current policy with buffered historical behaviours. Its main components are confidence-aware experience weighting, which assigns larger weights to low-confidence decisions, and decision-seeking behavior consolidation, which uses reverse KL divergence to encourage the current solver to preserve the high-probability actions recorded in previous behaviors. This makes LLR-BC particularly suitable for lifelong neural routing, as it consolidates decision patterns rather than entire models or regenerated instances. In our continual drifting scenarios, we adopt LLR-BC using the official implementation\footnote{https://github.com/PeiJY/LLR-BC} and the recommended hyperparameter settings. Since the original LLR-BC design caches experiences in the final epoch of each task, and each epoch in our continual drifting setting corresponds to a new task, we buffer experiences at every epoch following the same design principle. Appendix~\ref{app:LLRBC_less_buffering} further compares LLR-BC variants with different buffering frequencies, and the results show that this setting does not introduce negative performance effects. All other configurations are kept the same as in the original LLR-BC paper.

\paragraph{Li.}\textit{Li} (intra version) \cite{li2024enhancing} combines regularization with the experience replay strategy in the original framework. It periodically updates the exemplar model with a fixed epoch interval and regularizes the current policy toward the most recent exemplar model. The training objective combines the original reinforcement learning loss with a KL-divergence-based regularization term, weighted by a coefficient $\alpha$. This design encourages the model to retain desirable recent behaviours while continuing to adapt to the current problem distribution. In addition, the original framework revisits instances of previous tasks during learning new tasks by randomly selecting an old task and generating new instances of it on the fly.
We adapt Li to the continual drifting scenario with the updating interval as 25 epochs, identical to the original paper. Following the original formulation, the current policy is regularized toward the most recently saved reference model during training. Since the original paper does not specify a recommended value or tuning procedure for $\alpha$, we set $\alpha=0.5$, following~\cite{LLR-BC}.
Since new problem instances cannot be generated arbitrarily in our real-world continual drifting scenarios, we maintain an instance buffer with a size of 256 batches (comparable to the buffer size used by DREE) for each previously observed task using the instances encountered during training, maintained with reservoir sampling. 
All other settings are kept consistent with the original method whenever applicable. As no official implementation was available, we implemented it ourselves based on the methodological description in the paper.

\paragraph{MT-Ref.} MT-Ref is an intuitive multitask learning setting we included as a performance reference baseline. This method has access to all past and future tasks and can generate instances from each task on the fly. Specifically, for each batch in an epoch, it randomly selects one task and trains on newly generated instances from that task. Since all tasks remain jointly accessible throughout training, this reference does not suffer from catastrophic forgetting. Notably, \textit{MT-Ref} only serves as an idealized reference, and it is not practically applicable in lifelong learning scenarios. Thus, we only run MT-Ref once on task order 1 on the synthetic dataset, also because different task orders have the same principal tasks. In the real-world dataset, MT-Ref is run on the tasks of each region, as different regions have significantly different task characteristics. In addition, we design a more realistic cumulated multitask variant, which cannot access future tasks and instead incrementally records tasks as they arrive, denoted \textit{MT-cumu}. Every 200 epochs, it trains from scratch on all collected tasks using the same training procedure as the multitask reference for 200 epochs. The performance of this cumulated multitask variant is reported in Appendix~\ref{app:MTcumu}. The results show that DREE significantly outperforms this periodically retrained cumulated multitask method.

\paragraph{Excluded baselines.}
\cite{LLR-BC} involves more lifelong learning baselines, which are all model-buffering based, for comparison in the periodically stationary scenario, including Li (inter version), LiBOG~\cite{LiBOG}, Feng~\cite{feng2025lifelonglearner}, and EWC~\cite{EWC}. However, we do not compare against them in our continual drifting scenarios because these methods require storing a separate model for each stationary task and comparing the current model with every previously stored model during learning. In our setting, however, each epoch corresponds to a new task induced by the continuously changing real-world distribution. Therefore, the number of required historical models would grow with the number of epochs, making their computational and memory costs prohibitive. We also exclude \textit{Restart}, which trains an independent model for each task. Since each real-world task contains only one epoch of data, restarting from scratch for every task is unable to sufficiently learn the task-specific distribution and is expected to perform poorly. Furthermore, LLR-BC, which we have included, is verified to be able to outperform these methods~\cite{LLR-BC}. 

\section{Benchmark from Real-World Dataset}
\label{app:lade_benchmark}

\subsection{LaDe: Real-world Urban Delivery Dataset}
\label{app:dataset_stream_construction}

\paragraph{LaDe}
We obtain real-world daily delivery data from \textbf{LaDe} dataset \cite{LaDe}, which is a publicly available large-scale urban logistics dataset collected from real industrial operations.
LaDe covers five Chinese cities, namely \textit{Shanghai}, \textit{Hangzhou}, \textit{Chongqing}, \textit{Jilin}, and \textit{Yantai}, and spans six months from \textit{2022-05-01} to \textit{2022-10-31} \cite{LaDe}.
The dataset contains both pickup and delivery data. In this work, we focus on the package delivery data in LaDe, i.e., LaDe-Delivery. As the number of regions varies significantly between cities, we select one region per city, namely \texttt{sh-2}, \texttt{hz-120}, \texttt{jl-31}, \texttt{yt-167}, and \texttt{cq-144}, for equal analysis and experiments among cities.

LaDe-Delivery is organized as package-level delivery records rather than pre-defined combinatorial optimization instances.
Each record contains package-, courier-, spatial-, and temporal-related information, including package identifier, courier identifier, delivery date, geographic coordinates, city, region identifier, AOI information, and timestamps associated with the delivery process \cite{LaDe}.
Notably, LaDe-Delivery is not originally released as a benchmark for lifelong learning for VRP.
Instead, the original paper positions LaDe as a comprehensive real-world dataset for prediction problems like \emph{estimated time of arrival (ETA) prediction} and \emph{spatio-temporal graph forecasting} \cite{LaDe}.
% This distinction is important: our lifelong learning TSP benchmark is constructed from LaDe-Delivery, rather than directly provided by the original dataset.

Table~\ref{tab:region_characteristics} summarizes the key characteristics of the regions, including the number of valid days, the average original instance size, and the temporal coverage of each stream.
These regions differ substantially in scale and temporal evolution, providing diverse real-world lifelong learning environments.

\begin{table}[t]
\centering
\caption{
Characteristics of the selected regions. PpD for the number of packages per day.
}
\label{tab:region_characteristics}
\begin{tabular}{lccccc}
 \toprule
 Region & City  &  PpD Range &  PpD Mean (std.)\\
 \midrule
 sh-2   & Shanghai   & 2--175 & 56.9 (28.7) \\
 hz-120 & Hangzhou  & 88--608 & 327.8 (97.9)   \\
 jl-31  & Jilin   & 2--175 & 56.9 (28.7)  \\
 yt-167 & Yantai  & 9--108 & 42.2(21.8)  \\
 cq-144 & Chongqing   & 3--153 & 26.3 (32.0)   \\
 \bottomrule
\end{tabular}
\end{table}

\paragraph{From region-day package sets to a stream of real-world TSP instances.}
To transform LaDe-Delivery into a benchmark for lifelong combinatorial optimization, we group all package records by \emph{city}, \emph{region}, and \emph{delivery date}.
For each region-day tuple, all packages delivered in that region on that day are treated as nodes, and together form one raw TSP instance with real-world customers. Demand and capacity information are not provided in LaDe (e.g., the size/weight of a package), so we formulate TSP instances instead of CVRP instances. We do linear minmax normalization on each instance so that all nodes are located in $[0,1]^2$, following the common practice.
Formally, for a region $r$, and date $t$, we define
\[
I_{r,t} = \{x_1, x_2, \dots, x_{n_{r,t}}\},
\]
where each node $x_i$ corresponds to one package delivery location (coordinate) in the corresponding region-day package set, and $n_{r,t}$ is the number of packages on that day.

For a fixed region $r$, sorting these daily instances by date yields a chronological stream
\[
\mathcal{S}_{r} = \left(I_{r,1}, I_{r,2}, \dots, I_{r,T}\right),
\]
which we regard as a \textbf{stream of raw real-world TSP instances}.
At this stage, each instance is directly induced by a real operational day in a real delivery region.
% Importantly, this construction step only defines the stream itself; it does not yet involve instance resampling, train/test partitioning, or the lifelong evaluation protocol, which will be introduced later.

\begin{figure}[htbp]
    \centering

    % First row: three figures
    \begin{subfigure}{0.47\textwidth}
        \centering
        \includegraphics[width=\linewidth]{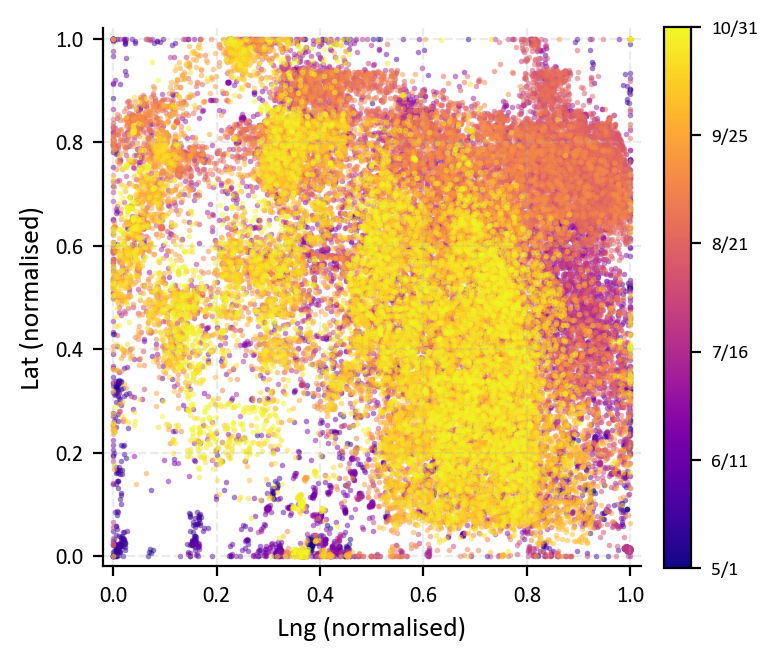}
        \caption{sh-2}
        \label{fig:sub1}
    \end{subfigure}
    \hfill
    \begin{subfigure}{0.47\textwidth}
        \centering
        \includegraphics[width=\linewidth]{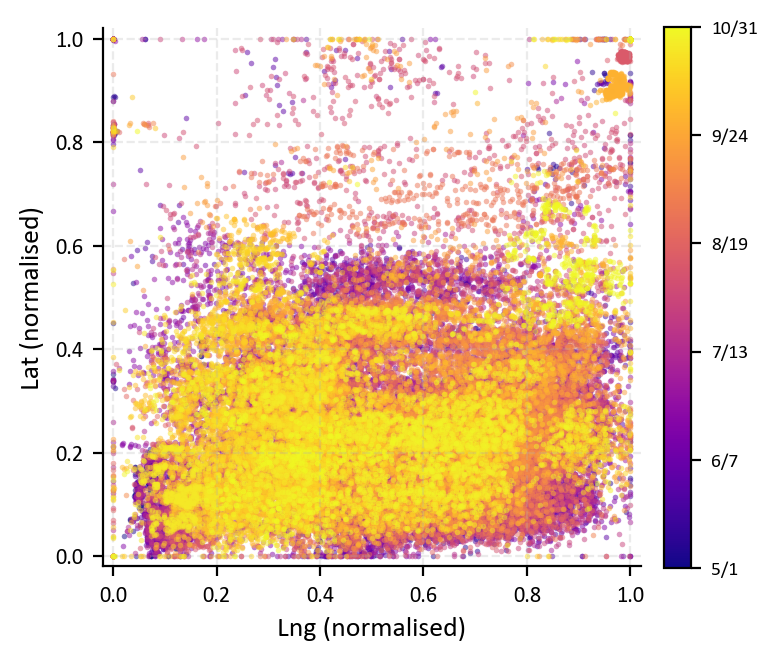}
        \caption{hz-120}
        \label{fig:sub2}
    \end{subfigure}

    % \vspace{0.3cm}
    
    \begin{subfigure}{0.47\textwidth}
        \centering
        \includegraphics[width=\linewidth]{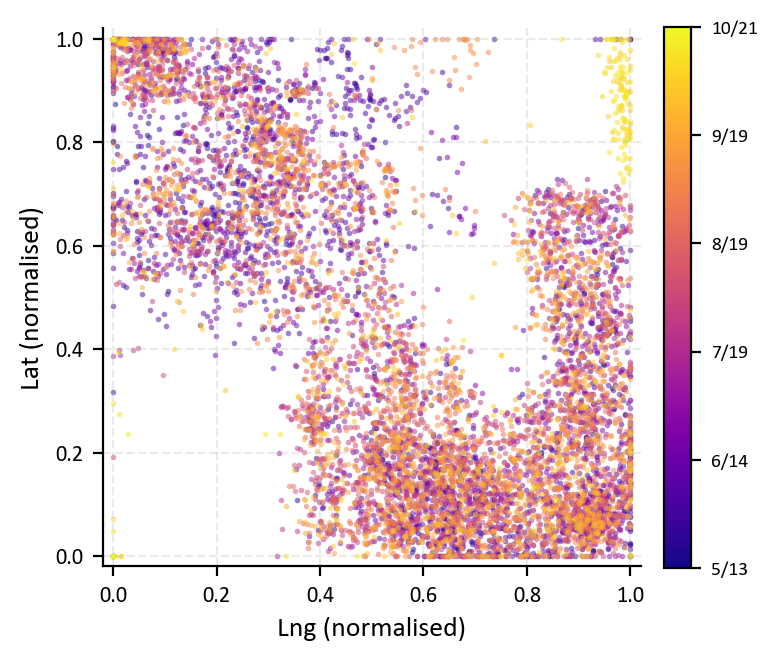}
        \caption{jl-31}
        \label{fig:sub3}
    \end{subfigure}
    \hfill
    % Second row: two figures, same size as above
    \begin{subfigure}{0.47\textwidth}
        \centering
        \includegraphics[width=\linewidth]{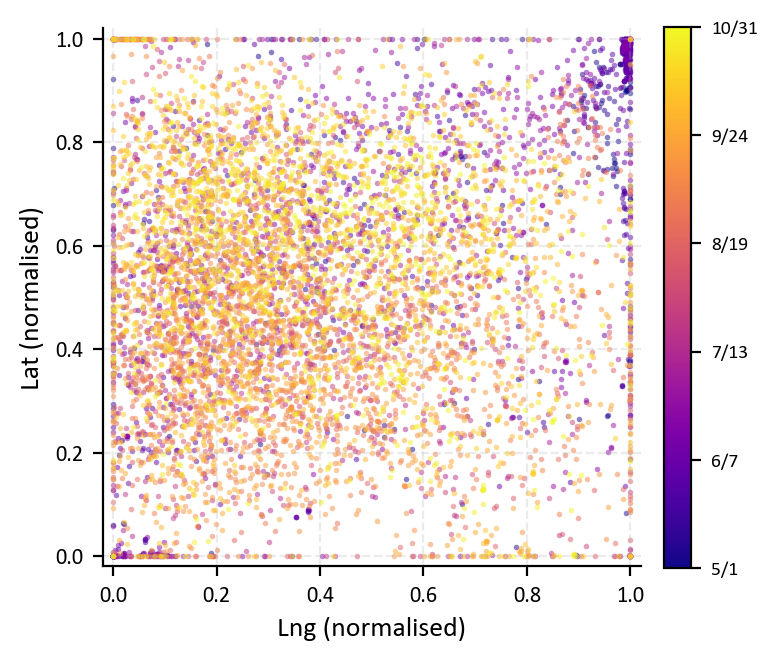}
        \caption{yt-167}
        \label{fig:sub4}
    \end{subfigure}
    % \hspace{0.08\textwidth}
    
    % \vspace{0.3cm}
    
    \begin{subfigure}{0.47\textwidth}
        \centering
        \includegraphics[width=\linewidth]{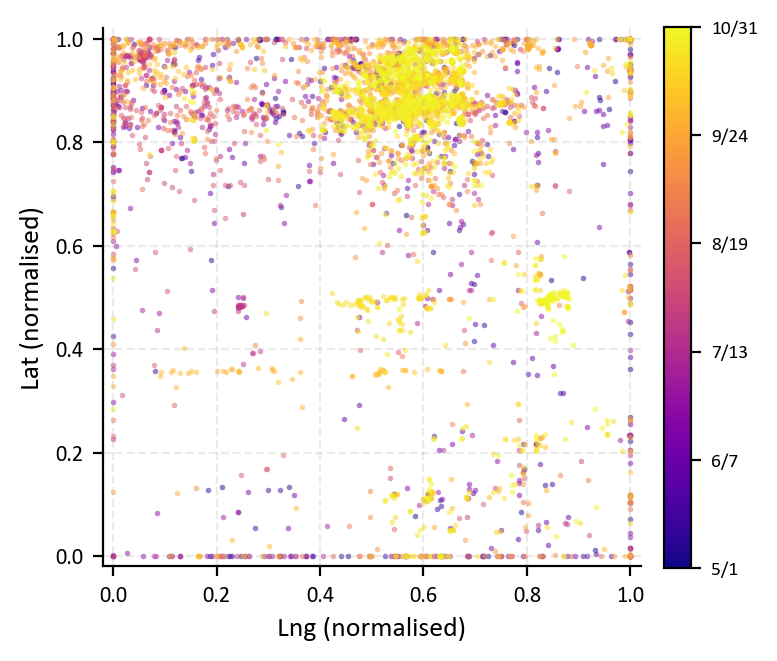}
        \caption{cq-144}
        \label{fig:sub5}
    \end{subfigure}

    \caption{TSP instances stream in each region. Nodes of different days are represented by different colors.}
    \label{fig:LaDe_scatter}
\end{figure}

Figure~\ref{fig:LaDe_scatter} demonstrates the location of nodes in each day and each region, with different days marked with different colors. Intuitively, some general drift trend is observed, like in the early days customers are more likely to concentrate in some clusters, and over the days, the customers gradually spread to the whole space.

\subsection{Real-World Continual Drift}
\label{app:real_world_drift}

Real-world delivery problem instance stream in each region exhibits \textbf{continual drift} over time.
We characterize such drift from two complementary perspectives: \textbf{problem scale} and \textbf{customer location distribution}.

\begin{figure}[htbp]
    \centering

    % First row: three figures
    \begin{subfigure}{0.47\textwidth}
        \centering
        \includegraphics[width=\linewidth]{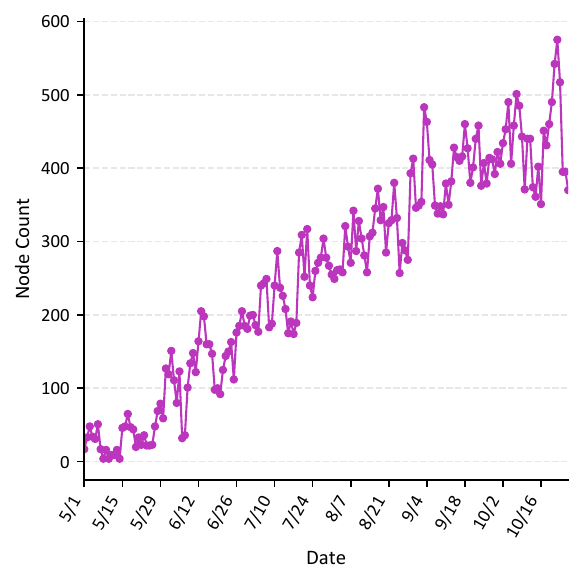}
        \caption{sh-2}
        \label{fig:sub1}
    \end{subfigure}
    \hfill
    \begin{subfigure}{0.47\textwidth}
        \centering
        \includegraphics[width=\linewidth]{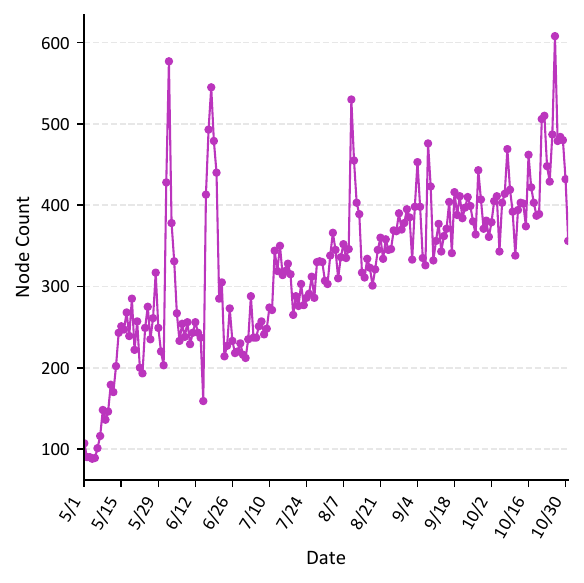}
        \caption{hz-120}
        \label{fig:sub2}
    \end{subfigure}

    % \vspace{0.3cm}
    
    \begin{subfigure}{0.47\textwidth}
        \centering
        \includegraphics[width=\linewidth]{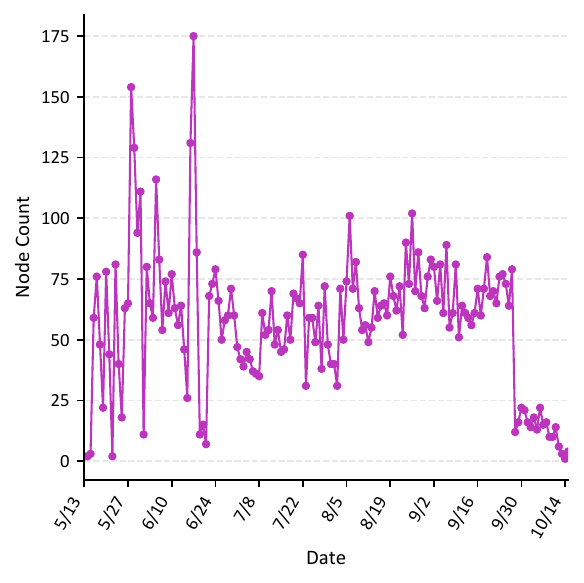}
        \caption{jl-31}
        \label{fig:sub3}
    \end{subfigure}
    \hfill
    % Second row: two figures, same size as above
    \begin{subfigure}{0.47\textwidth}
        \centering
        \includegraphics[width=\linewidth]{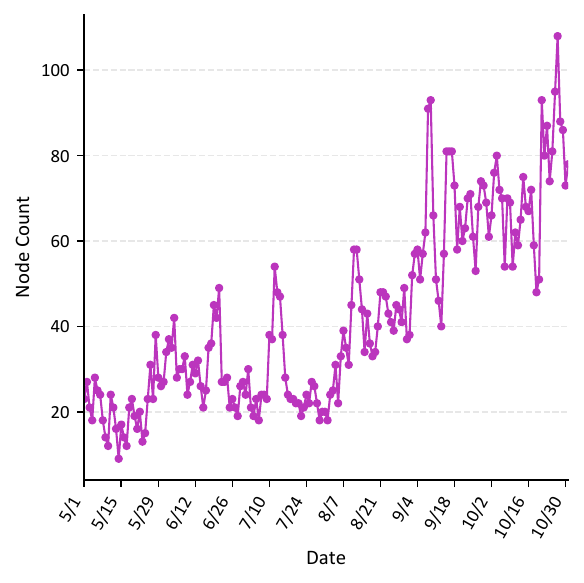}
        \caption{yt-167}
        \label{fig:sub4}
    \end{subfigure}
    % \hspace{0.08\textwidth}
    
    % \vspace{0.3cm}
    
    \begin{subfigure}{0.47\textwidth}
        \centering
        \includegraphics[width=\linewidth]{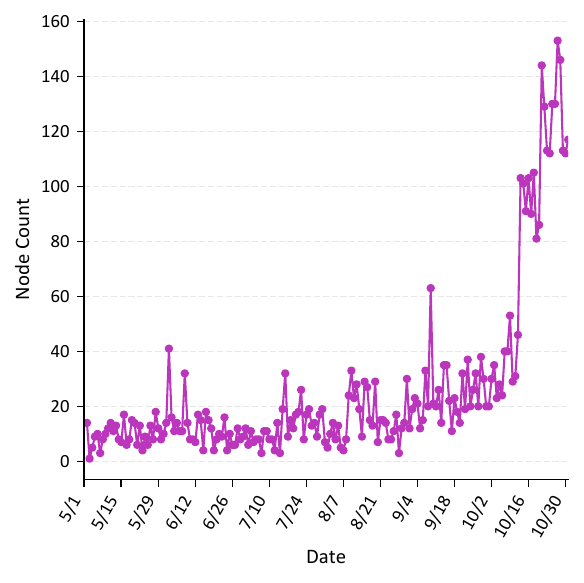}
        \caption{cq-144}
        \label{fig:sub5}
    \end{subfigure}

    \caption{Daily instance sizes over time for the selected region streams. Although local fluctuations exist, the overall scale changes progressively in most of the cases, indicating real-world continual drift in problem size.}
    \label{fig:LaDe_scale_curve}
\end{figure}

\paragraph{Scale drift.}
For each region stream $\mathcal{S}_{r}$, we first track the daily instance size $n_{r,t}$, i.e., the number of package nodes in the corresponding region-day instance.
By visualizing $n_{r,t}$ over time (cf. Figure~\ref{fig:LaDe_scale_curve}), we observe that although local fluctuations exist, the overall trend shows a pattern of continual drift. 
This indicates that the real-world stream exhibits gradual temporal variation in instance size.

\paragraph{Distribution drift.}
The problem scale alone does not fully characterize the drifting of the delivery distribution.
We therefore further quantify the structural discrepancy between instances using the \textbf{Gromov--Wasserstein (GW) distance} \cite{GWdistance}, which is suitable for comparing relational structures defined on different node sets.

Given two TSP instances $I_t$ and $I_{t+h}$ with empirical node distributions $\mu_t$ and $\mu_{t+h}$ and intra-instance node matrices $D_t \in \mathbb{R}^{n_t \times n_t}$ and $D_{t+h} \in \mathbb{R}^{n_{t+h} \times n_{t+h}}$, respectively, we compute their GW distance as
\[
\mathrm{GW}(I_t, I_{t+h})
=
\min_{\pi \in \Pi(\mu_t,\mu_{t+h})}
\left(
\sum_{i,i',j,j'}
\left|D_t(i,i') - D_{t+h}(j,j')\right|^2
\pi_{ij}\pi_{i'j'}
\right)^{1/2},
\]
where $\Pi(\mu_t,\mu_{t+h})$ denotes the set of couplings between the two empirical measures.
In our implementation, the ground structure (i.e., $D$) of each instance is defined by its pairwise Euclidean distance matrix over node coordinates.

To measure temporal drift at different scales, we compute the \textbf{average lag-$h$ GW distance}:
\[
\overline{\mathrm{GW}}(h)
=
\frac{1}{T-h}
\sum_{t=1}^{T-h}
\mathrm{GW}(I_t, I_{t+h}),
\qquad h = 1,2,\dots,H.
\]
If the stream is gradually drifting, then instances that are farther apart in time should, on average, become more dissimilar.
Accordingly, we expect $\overline{\mathrm{GW}}(h)$ to show an increasing trend as the lag $h$ grows, possibly with local irregularities due to short-term operational noise.

\begin{figure}[htbp]
    \centering

    % First row: three figures
    \begin{subfigure}{0.47\textwidth}
        \centering
        \includegraphics[width=\linewidth]{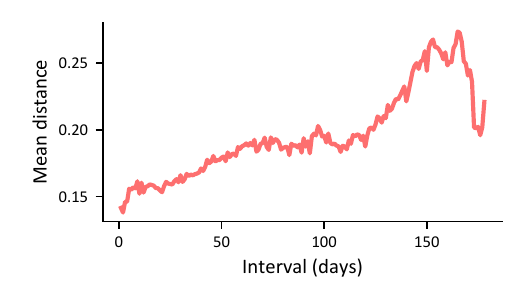}
        \caption{sh-2}
        \label{fig:sub1}
    \end{subfigure}
    \hfill
    \begin{subfigure}{0.47\textwidth}
        \centering
        \includegraphics[width=\linewidth]{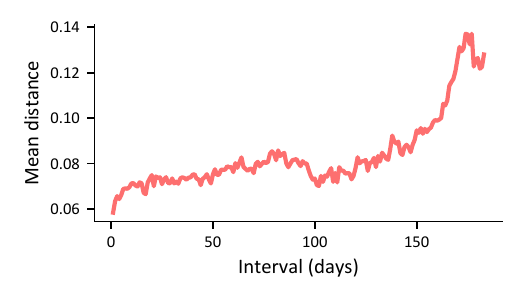}
        \caption{hz-120}
        \label{fig:sub2}
    \end{subfigure}

    % \vspace{0.3cm}
    
    \begin{subfigure}{0.47\textwidth}
        \centering
        \includegraphics[width=\linewidth]{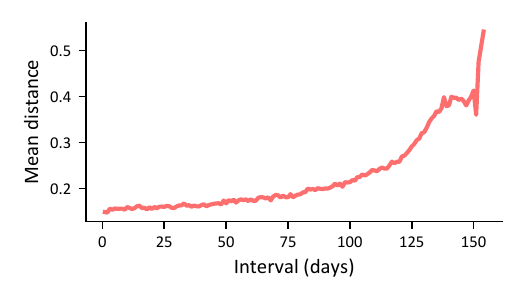}
        \caption{jl-31}
        \label{fig:sub3}
    \end{subfigure}
    \hfill
    % Second row: two figures, same size as above
    \begin{subfigure}{0.47\textwidth}
        \centering
        \includegraphics[width=\linewidth]{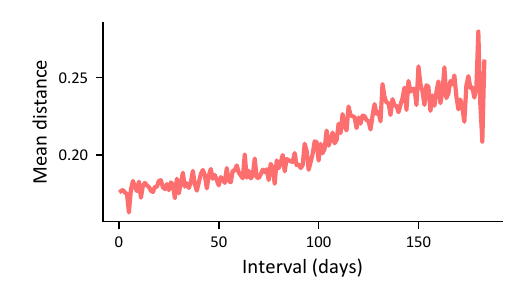}
        \caption{yt-167}
        \label{fig:sub4}
    \end{subfigure}
    % \hspace{0.08\textwidth}
    
    % \vspace{0.3cm}
    
    \begin{subfigure}{0.47\textwidth}
        \centering
        \includegraphics[width=\linewidth]{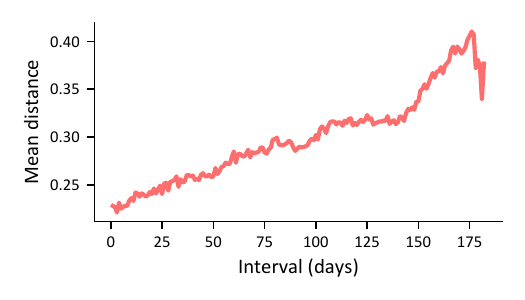}
        \caption{cq-144}
    \end{subfigure}

    \caption{Average lag-$h$ GW distance for the selected region streams. The average structural discrepancy generally increases with temporal lag, supporting the existence of continual distribution drift in real-world delivery problems.}
    \label{fig:LaDe_gs_curve}
\end{figure}

Figure~\ref{fig:LaDe_gs_curve} reports the average lag-$h$ GW distance curves. The downturn at the end of the curves suggests that previous patterns, or patterns similar to earlier ones, may reappear over time. This highlights the importance of retaining knowledge from previous tasks.
We observe that the average structural discrepancy generally increases with temporal lag, again with local fluctuations. 
Together, these two analyses provide direct evidence that the real-world delivery streams exhibit the target form of \textbf{continual drift}: the underlying problem distribution changes over time in a globally progressive but locally noisy manner.

\subsection{From Raw TSP Instance Streams to a Real-World Lifelong Learning Dataset\label{app:lade_lifelong_scenario}}

We next convert the region-day-level real-world TSP streams into a lifelong learning scenario for benchmarking continual combinatorial optimization methods.
For each selected region, we retain only the days whose original region-day instance contains at least 10 package nodes, in order to avoid extremely small and uninformative cases.

\paragraph{Instance resampling.}
The original real-world data stream contains large-scale instances with nearly one thousand customers, which exceed the capability of the base solvers used in this work. However, addressing such large-scale optimization challenges is beyond the primary scope of this study.
To obtain solver-compatible instances while preserving the original spatial distribution, we generate benchmark instances by \textbf{node resampling} from each original region-day instance.
Specifically, for an original instance with $n_t$ nodes, we sample a subset whose size is determined by a region-specific ratio $q$ and then clip it into the interval $[10,100]$:
\[
\tilde{n}_t = \mathrm{clip}\left(\left\lfloor q \cdot n_t \right\rfloor,\, 10,\, 100\right).
\]
The value of $q$ is chosen separately for different regions according to their original scale characteristics. $q$ is 0.3, 0.2, 0.6, 0.9 and 0.7 for sh-2, hz-120, jl-31, yt-167 and cq-144, respectively.
The same resampling protocol is applied to all benchmark instances used in training and testing.

Another benefit of resampling is that, for each time step, namely each day, we obtain a distribution of problem instances rather than a single problem instance. This is consistent with the design principle of existing neural solvers, which typically learn from multiple problem instances sampled from an underlying distribution rather than from one isolated example.

\paragraph{Training and test splits.}
For each region stream, we sort the valid daily instances chronologically.
The first {95\% of days} are used to form the lifelong training stream, while the last {5\% of days} are held out as a future generalization test set, for obtaining Gen value.
This split evaluates whether a lifelong method can generalize to future real-world data beyond the training horizon.

In addition, to measure adaptation and forgetting on previously encountered patterns, we construct a process test set, similar to the principal-task test set in the synthetic dataset, for obtaining AP, AFB, AMFB and ABPl.
Concretely, we generate one test instance set every {30 days} along the training horizon.
These periodic test sets are built from the corresponding historical real-world instances and are used to evaluate the performance of each lifelong learning method on previously observed tasks/patterns over time.

% Therefore, our benchmark contains two types of test sets:
% \begin{itemize}
% \item \textbf{Periodic retrospective test sets}: used to evaluate adaptation and forgetting on previously encountered patterns.
% \item \textbf{Future generalization test set}: used to evaluate forward generalization on the held-out tail split.
% \end{itemize}

\paragraph{Optimal solutions and gap computation.}
Following the setting on the synthetic dataset, we use Gurobi to solve each test instance in this real-world dataset to obtain an optimal solution whenever tractable.
These optimal solutions are used to compute the optimality gap of each compared lifelong learning method.
For a method producing solution length $L_i$ on instance $i$, with Gurobi optimum $L_i^\star$, we define the relative optimality gap as
\[
\mathrm{Gap}_i = \frac{L_i - L_i^\star}{L_i^\star}\times 100\%.
\]
We report the average gap over each test set.
The Gurobi-derived optima also serve as the reference for evaluating the absolute solution quality of different lifelong learning methods.

\section{Details of Synthetic Dataset\label{app:synthetic_dataset}}

\subsection{Lifelong Learning Tasks\label{app:synthetic_task}}

We adopt the six tasks ($P^U,P^R,P^{GM},P^E,P^C,P^G$) from~\cite{LLR-BC} as our principal tasks for the synthetic dataset and restate the instance-generation procedure here for completeness.

The six tasks are named with the node location distribution: Uniform (U), Gaussian Mixture (GM), Explosion (E), Compression (C), Grid (G), and Ring (R). Four of them follow the coordinate-generation protocols used in~\cite{evolveTSP}, while the Grid and Ring tasks are additionally included in \cite{LLR-BC} to increase the diversity of spatial structures. Tasks U and R correspond to scale 20, tasks G and E to scale 50, and tasks C and GM to scale 100, following \cite{LLR-BC}. For CVRP, each task is also associated with a task-specific demand distribution. Unless otherwise specified, customer/city coordinates are normalized to the unit square $[0,1]^2$. For CVRP, the depot location is sampled independently from $\mathcal{U}([0,1]^2)$. Following~\cite{POMO}, the vehicle capacity is set to $30$, $40$, and $50$ for problem sizes $20$, $50$, and $100$, respectively.

\paragraph{Uniform (U).}
For each customer/city node $i$, its coordinate is sampled independently as
\[
(x_i,y_i) \sim \mathcal{U}([0,1]^2).
\]
For CVRP, the customer demand is sampled from a discrete uniform distribution
over $\{1,\ldots,10\}$.

\begin{figure}[h]
\centering
\begin{subfigure}{.4\linewidth}
 \centering
 \includegraphics[width=\linewidth]{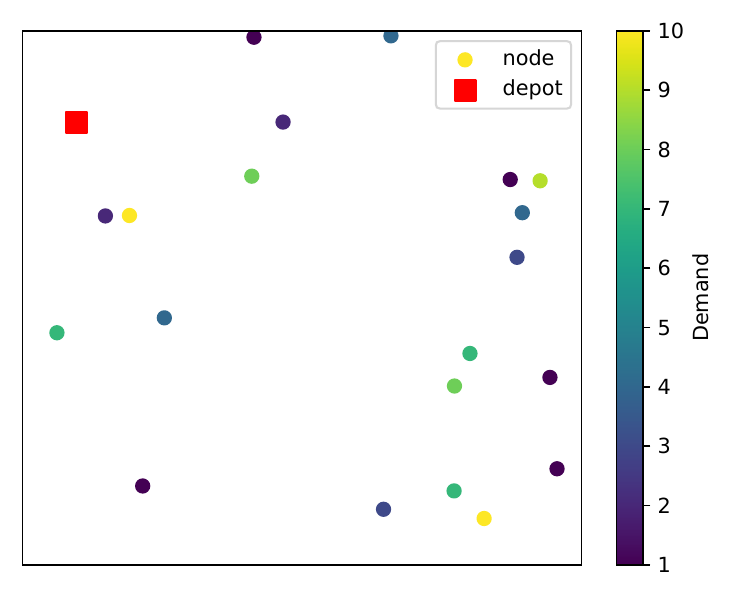}
 \caption{CVRP}
\end{subfigure}
\begin{subfigure}{.318\linewidth}
 \centering
 \includegraphics[width=\linewidth]{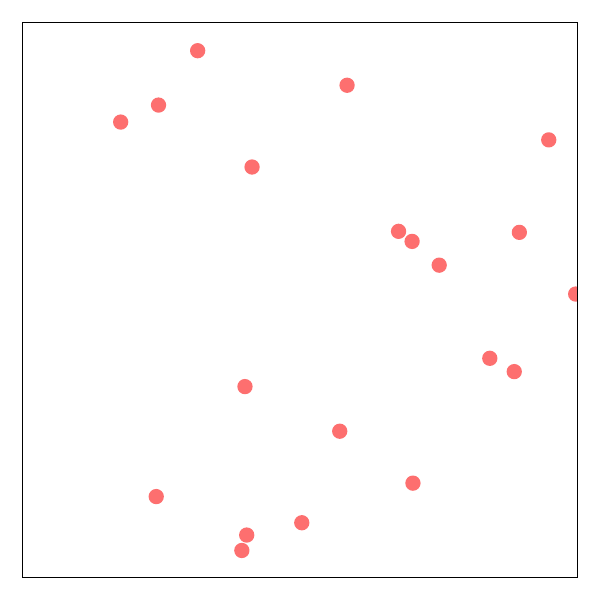}
 \caption{TSP}
\end{subfigure}
\caption{Example problem instance of Task $P^U$}
\end{figure}

\paragraph{Gaussian Mixture (GM).}
We first sample five cluster centers $\{c_z\}_{z=1}^{5}$ independently from
$[0,50]^2$. Around each center, customer/city nodes are generated from an
isotropic Gaussian distribution,
\[
(x_i,y_i) \sim \mathcal{N}(c_z, I),
\]
where $I$ denotes the identity covariance matrix. The resulting coordinates are
then mapped to $[0,1]^2$ using min--max normalization. For CVRP, the demand is
made dependent on the local cluster structure. Specifically, for each node $i$,
we compute its Euclidean distance to the corresponding cluster center, normalize
these distances to $[0,1]$, and set the demand to the rounded value of
$10 \cdot \mathrm{dist}_i$.

\begin{figure}[h]
\centering
\begin{subfigure}{.4\linewidth}
 \centering
 \includegraphics[width=\linewidth]{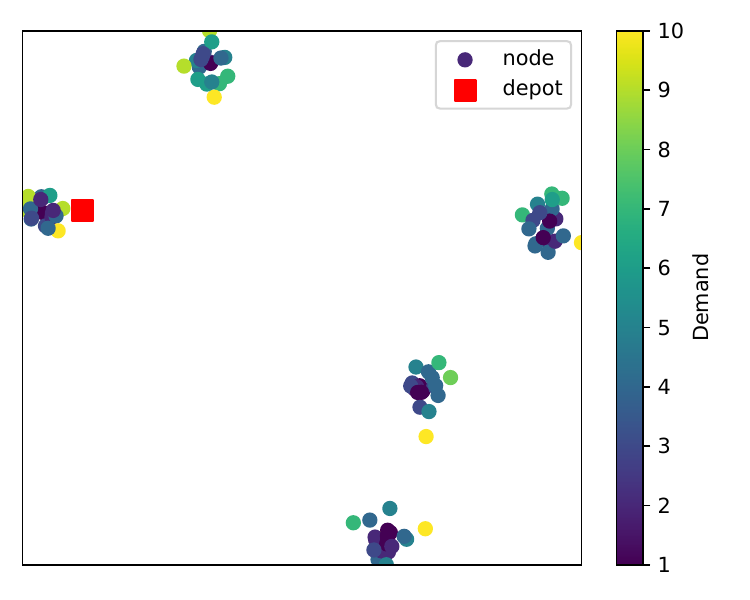}
 \caption{CVRP}
\end{subfigure}
\begin{subfigure}{.318\linewidth}
 \centering
 \includegraphics[width=\linewidth]{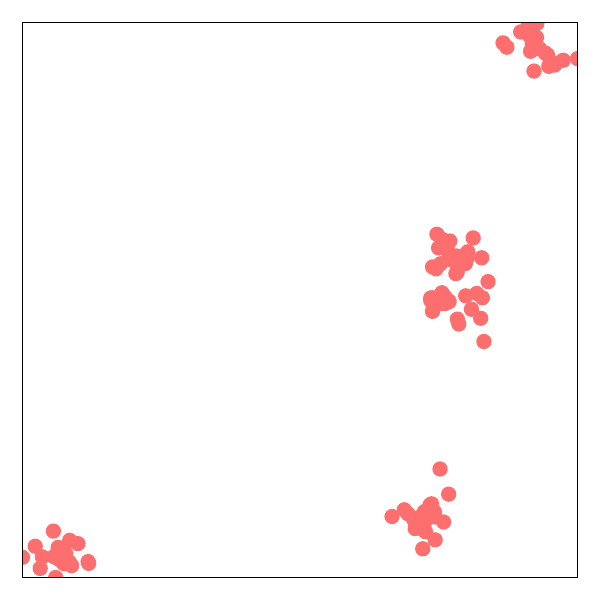}
 \caption{TSP}
\end{subfigure}
\caption{Example problem instance of Task $P^{GM}$}
\end{figure}

\paragraph{Explosion (E).}
The initial node coordinates are sampled uniformly from $[0,1]^2$. A reference
point $p \sim \mathcal{U}([0,1]^2)$ is then sampled independently. For every node
whose distance to $p$ is smaller than $0.3$, the node is displaced away from
$p$ by a distance $0.3+s$, where $s \sim \mathrm{Exp}(40)$. After the
displacement, all coordinates are clipped to $[0,1]^2$. For CVRP, each demand is
sampled from $\mathcal{N}(5,1)$, rounded to the nearest integer, and clipped to
the range $[0,10]$.

\begin{figure}[h]
\centering
\begin{subfigure}{.4\linewidth}
 \centering
 \includegraphics[width=\linewidth]{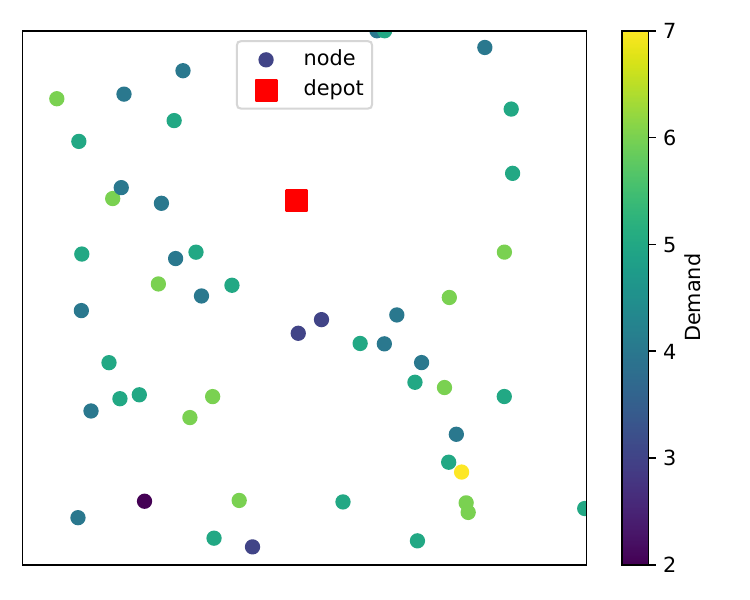}
 \caption{CVRP}
\end{subfigure}
\begin{subfigure}{.318\linewidth}
 \centering
 \includegraphics[width=\linewidth]{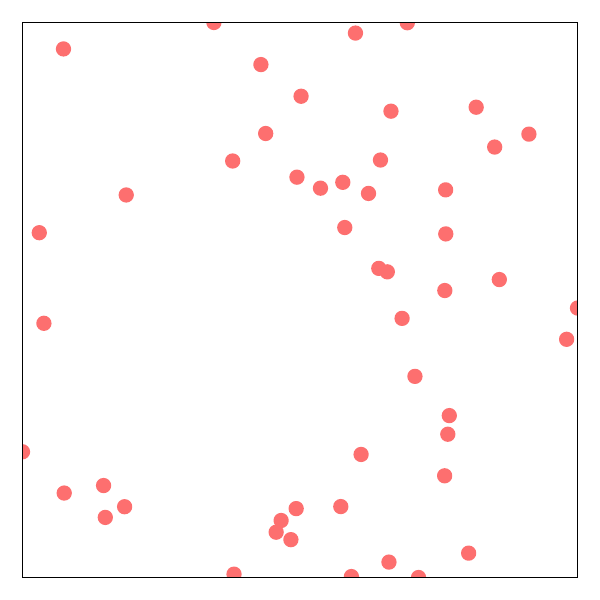}
 \caption{TSP}
\end{subfigure}
\caption{Example problem instance of Task $P^E$}
\end{figure}

\paragraph{Compression (C).}
Nodes are first sampled uniformly from $[0,1]^2$. We then sample two points
$p_1,p_2 \sim \mathcal{U}([0,1]^2)$ and use them to define a line $l$. For each
node whose distance to $l$ is below $0.3$, a new signed distance is sampled from
$\mathcal{N}(0,0.1^2)$, and the node is moved along the direction perpendicular
to $l$ so that its new distance to the line matches the sampled value. The final
coordinates are clipped to $[0,1]^2$. For CVRP, the demand is generated as
$10-x$, where $x \sim \mathcal{N}(5,1)$, followed by rounding and clipping to
$[0,10]$.

\begin{figure}[h]
\centering
\begin{subfigure}{.4\linewidth}
 \centering
 \includegraphics[width=\linewidth]{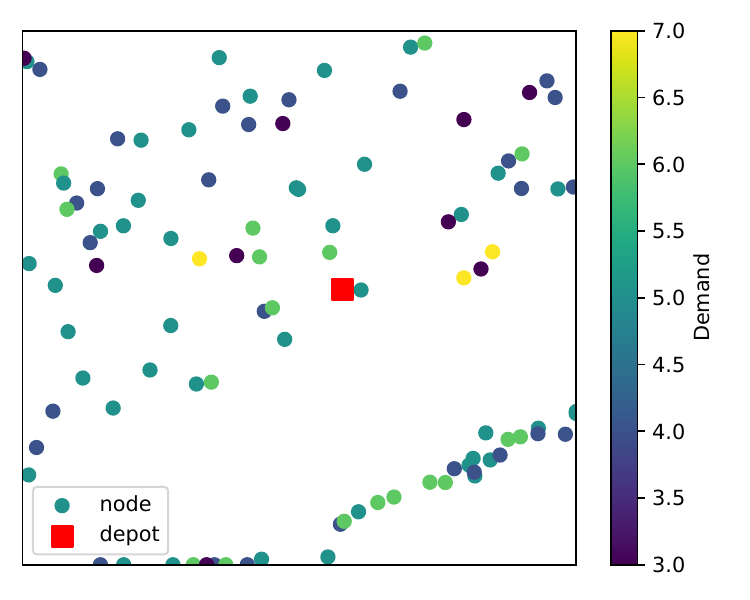}
 \caption{CVRP}
\end{subfigure}
\begin{subfigure}{.318\linewidth}
 \centering
 \includegraphics[width=\linewidth]{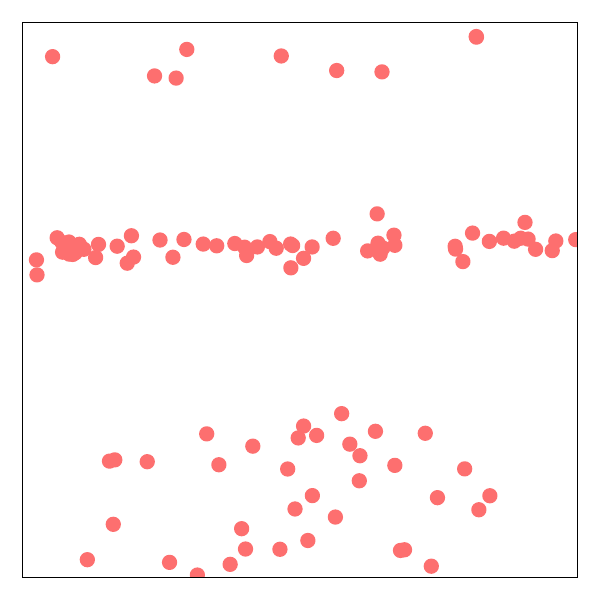}
 \caption{TSP}
\end{subfigure}
\caption{Example problem instance of Task $P^C$}
\end{figure}

\paragraph{Grid (G).}
We first sample an aspect ratio parameter $\rho \sim \mathcal{U}(0.2,0.8)$.
With probability $0.5$, we set the rectangle width and height to
$(w,h)=(1,\rho)$; otherwise, we set $(w,h)=(\rho,1)$. The center of the
rectangle is sampled as
\[
c_x \sim \mathcal{U}\left(\frac{w}{2}, 1-\frac{w}{2}\right), \qquad
c_y \sim \mathcal{U}\left(\frac{h}{2}, 1-\frac{h}{2}\right).
\]
A grid is then constructed inside this rectangle. Following the original
benchmark protocol~\cite{LLR-BC}, the number of grid cells along the
$x$- and $y$-axes is set to
\[
a=\left\lceil \sqrt{n\frac{w}{h}} \right\rceil, \qquad
b=\left\lceil \sqrt{\frac{n}{a}} \right\rceil .
\]
Nodes are placed into the grid cells until the required number of customers is
reached. If the grid cannot be evenly filled, the remaining empty cells are
chosen from the largest-$x$ cells in the row with the largest $y$ value. For
CVRP, the demand of each customer is computed from its depot distance with an
additive noise sampled from $\mathcal{U}(0,1)$. The noisy distances are linearly
mapped to $[1,10]$ and then rounded.

\begin{figure}[h]
\centering
\begin{subfigure}{.4\linewidth}
 \centering
 \includegraphics[width=\linewidth]{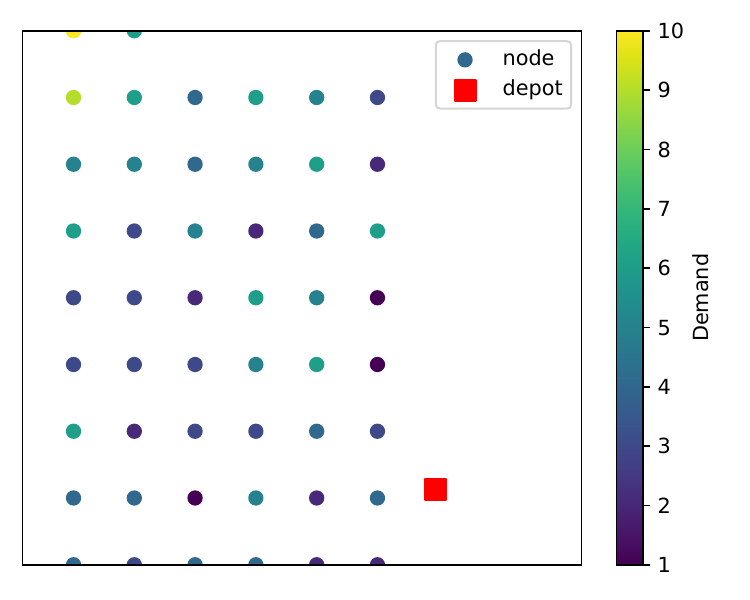}
 \caption{CVRP}
\end{subfigure}
\begin{subfigure}{.318\linewidth}
 \centering
 \includegraphics[width=\linewidth]{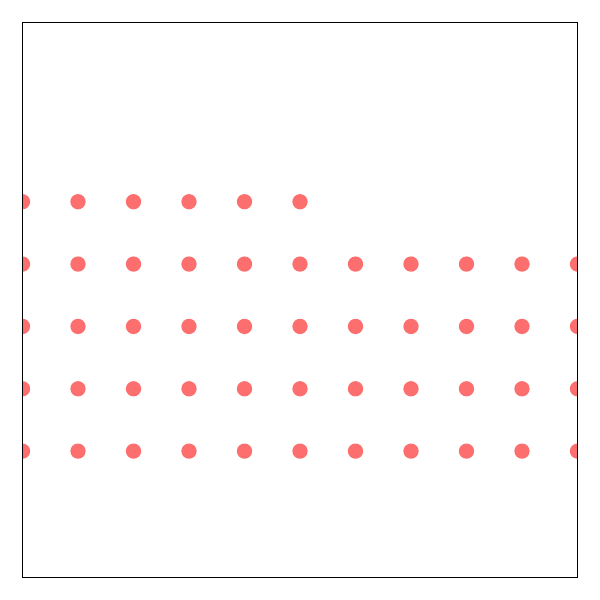}
 \caption{TSP}
\end{subfigure}
\caption{Example problem instance of Task $P^G$}
\end{figure}

\paragraph{Ring (R).}
For each node, we sample an angle $\theta \sim \mathcal{U}(0,2\pi)$ and a radius
\[
r = r_1 + r_2, \qquad
r_1 \sim \mathcal{U}(0.3,0.4), \quad
r_2 \sim \mathcal{N}(0,0.05^2).
\]
The preliminary coordinate is then given by
\[
(x,y) = (0.5 + r\cos\theta,\; 0.5 + r\sin\theta).
\]
Following~\cite{LLR-BC}, an additional anisotropic scaling is applied to
one coordinate axis using a random scaling factor, producing horizontally or
vertically compressed ring-like structures. For CVRP, customer demands are
derived from the depot distance with additive noise sampled from
$\mathcal{U}(0,2)$. The noisy distances are mapped to $[1,10]$ and rounded.

\begin{figure}[h]
\centering
\begin{subfigure}{.4\linewidth}
 \centering
 \includegraphics[width=\linewidth]{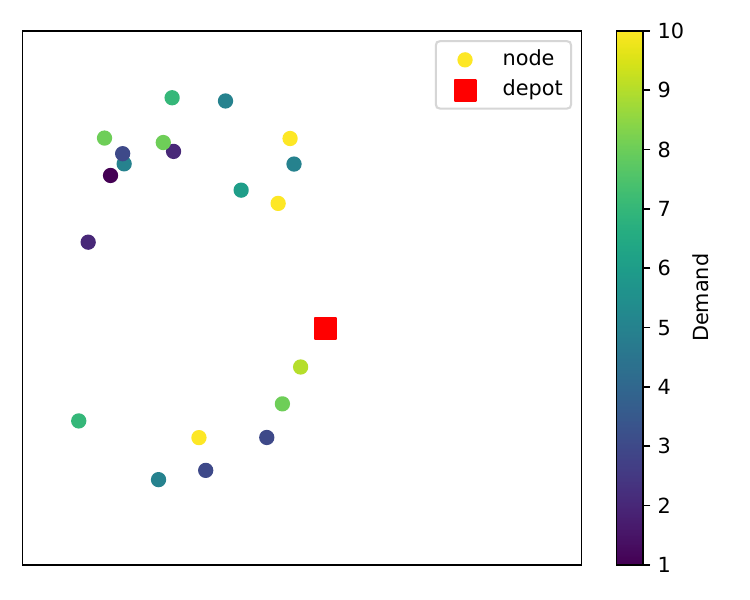}
 \caption{CVRP}
\end{subfigure}
\begin{subfigure}{.318\linewidth}
 \centering
 \includegraphics[width=\linewidth]{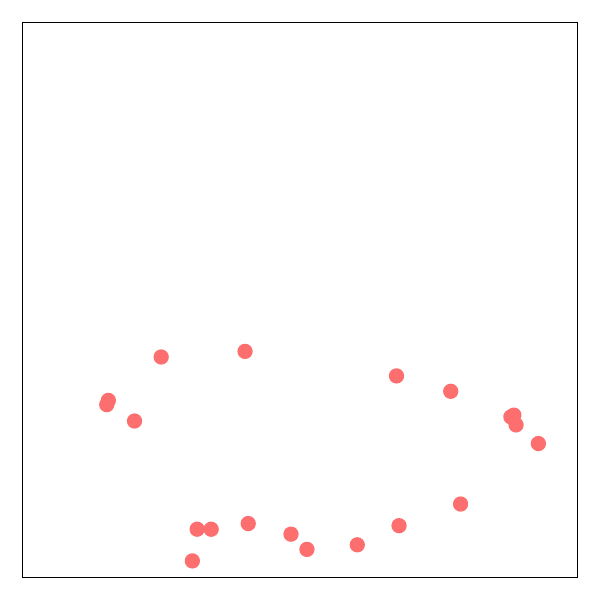}
 \caption{TSP}
\end{subfigure}
\caption{Example problem instance of Task $P^R$}
\end{figure}

\paragraph{Task Orders.} For both TSP and CVRP, we use the five random task orders from~\cite{LLR-BC}, as follows. 

\begin{itemize}
\item Order 1: E $\rightarrow$ C $\rightarrow$ G $\rightarrow$ U $\rightarrow$ R $\rightarrow$ GM.
\item Order 2: U $\rightarrow$ GM $\rightarrow$ E $\rightarrow$ R $\rightarrow$ G $\rightarrow$ C.
\item Order 3: E $\rightarrow$ G $\rightarrow$ R $\rightarrow$ C $\rightarrow$ U $\rightarrow$ GM.
\item Order 4: G $\rightarrow$ GM $\rightarrow$ E $\rightarrow$ U $\rightarrow$ R $\rightarrow$ C.
\item Order 5: G $\rightarrow$ C $\rightarrow$ R $\rightarrow$ U $\rightarrow$ GM $\rightarrow$ E.
\end{itemize}

\paragraph{Task Sequences with Continual Drift}
In the continually drifting scenario, we use an epoch as the time step unit. Drift occurs only between epochs, and the task is assumed to be stationary within each epoch. To construct such a stream of tasks, we linearly mix up principal tasks. 
We assign $K=6$ predefined principal tasks $\{P^i\}_{i=1}^K$ evenly to $K$ epochs with interval of $m=\frac{T}{K-1}=\frac{1000}{6-1}=200$ epochs, i.e., to epochs 0, $m $, $2m$, $\dots$, $T$. The task of an epoch between the epochs of two principal tasks is a linear mixture of them. Specifically, an epoch $t \in \{im+1,\dots,(i+1)m-1\}$, we generate an intermediate task $P_{t}$ with scale $S_{t} = \left\lceil \frac{(i+1)m-t}{m}S^{i} + \frac{t-im}{m} S^{i+1} \right\rfloor$, with $S^{i}$ and $S^{i+1}$ as the problem scales of $P^i$ and $P^{i+1}$, respectively. For an instance of $P_{t}$, $\left\lceil \frac{t-im}{m} S_t\right\rfloor$ nodes are generated with the distribution of $P^{i+1}$ while the rest are generated with the distribution of $P^{i}$.

Notably, problem scale and node coordinate distribution change between tasks for both CVRP and TSP. For CVRP, the node demand will also change. The coordinate of the depot for CVRP remains unchanged for all tasks.

Since VRP instances are discrete and combinatorial, whose natural unit of change is a node, epoch-level granularity is sufficient for our continual drift design. Specifically, with principal problem scales of 20, 50, and 100 and 200 epochs between consecutive principal tasks, a change of one node in scale (or, equivalently, a one-node deviation from a principal distribution) corresponds to roughly two epochs. Hence, treating each epoch as a time step provides adequate resolution, and using finer-grained units would not significantly affect the outcomes.

\subsection{Training and Test Settings~\label{app:synthetic_tt_setting}}

Following~\cite{LLR-BC}, we assume the problem instance generation is uncontrollable.
For all methods, during training, we use data augmentation with an augmentation factor of 8, following common settings~\cite{POMO,INViT}. During testing, augmentation is not used.

With 1000 epochs and 128 batches per epoch, DREE with $|\mathcal{B}| = 256$ can retain only about 0.2\% of the instances generated throughout lifelong learning. Nevertheless, given its strong performance, we consider this cost acceptable.

We conduct tests only on the six principal tasks. This choice is necessary because the intermediate tasks induced by continual drift can vary across different task orders, whereas the six principal tasks always appear regardless of the order. Testing these shared tasks, therefore, enables a fair and comprehensive comparison across orders. Since the principal tasks are uniformly distributed over the learning horizon, performance on them is also representative of overall lifelong-learning behavior. We also conduct an additional comparison on task order 1 with finer-grained test set generation, and the results lead to the same conclusion on the six-principal-task test setting (cf. Appendix~\ref{app:Fine-grained_test}), suggesting the effectiveness and efficiency of the setting.

For testing, we report the optimality gap. For each test instance, we obtain an optimal solution using HGS for CVRP and Gurobi for TSP, compute the gap per instance, and then average gaps over the test set. All task orders share the same test-instance sets. As for each task order there are 1000 possible models from the 1000 epochs to test, and each of them need to be test on all six principal tasks, it would be very expensive to test all of them. Thus, we select the model for every 20 epoch to test, which still give a comprehensive description of solvers' performance changing during lifelong learning process.

DREE trains on 128 batches of new-task instances per epoch. We record the number of PIR steps and find that, across different task orders, it is approximately 36 (avg. $N = \frac{128}{36}\simeq 3.56$ ). Therefore, for other methods (i.e., \textit{fine-tuning},
% \textit{MT-Ref},
LLR-BC, and \textit{Li}), we run $128+36=164$ batches per epoch so that they execute the same number of solving episodes, while processing even more training instances. This ensures that DREE’s performance gains are not attributable to longer training time or more training instances.

Following~\cite{LLR-BC}, we use representative subsets of TSPLIB and CVRPLIB (Set-X), as listed in Tables~\ref{tab:cvrplib} and Table~\ref{tab:tsplib}. 

\section{More Experiment Results on Real-world Dataset \label{app:LaDe_result_details}}

Table~\ref{tab:detailed_metric_LaDe} reports the detailed performance of all methods on each region. Here, \textit{Gen} denotes the zero-shot generalization performance on the held-out instances from the last 5\% of valid days after training on the instances from the first 95\% of days. The results show that DREE consistently achieves the best lifelong learning performance across the task streams from all five regions, obtaining the highest AP values in all regions, including three cases where it even surpasses MT-Ref, and the highest APl values in all regions, including four cases where it outperforms MT-Ref. Moreover, DREE also exhibits the strongest overall generalization ability for solving potential future problem instances, outperforming all other lifelong learning methods on four out of the five regions. These results demonstrate the strong effectiveness of DREE in learning from real-world routing scenarios with drifts.

\begin{table*}[h]

\centering
\caption{Detailed metric values on the real-world dataset. \textit{Gen} represents the average test performance on test instances from the held-out final 5\% of days.\label{tab:detailed_metric_LaDe}}
\begin{tabular}{c|c|c|c|c|c|c}
\toprule
Region& {Method} & {AP} & {AFB} & {AMFB} & {ABPl} & {Gen} \\
\midrule
\multirow{5}{*}{sh-2}& MT-Ref  & 0.59  & -  & -  & - & 0.85  \\ \cmidrule(lr){2-7} 
& FT& 1.83  & 1.56  & 4.38  & 0.27 & 0.82  \\
& Li& 9.11  & 0.23  & 1.75  & 8.88 & 11.18 \\
& LLR-BC  & 0.59  & 0.48  & 0.75  & 0.11 & 0.43  \\
& \textbf{DREE} & 0.64  & 0.41  & 0.68  & 0.23 & 0.79  \\ \midrule
\multirow{5}{*}{hz-120} & MT-Ref  & 1.64  & -  & -  & - & 1.68  \\ \cmidrule(lr){2-7} 
& FT& 0.32  & 0.2& 0.78  & 0.12 & 0.28  \\
& Li& 3.75  & 0.37  & 0.49  & 3.38 & 5.42  \\
& LLR-BC  & 0.46  & 0.34  & 0.63  & 0.11 & 0.37  \\
& \textbf{DREE} & 0.27  & 0.21  & 0.34  & 0.06 & 0.4\\ \midrule
\multirow{5}{*}{jl-31}  & MT-Ref  & 0.54  & -  & -  & - & 2.11  \\ \cmidrule(lr){2-7} 
& FT& 2.71  & 2.15  & 5.03  & 0.55 & 2.45  \\
& Li& 12.4  & 7.46  & 15.86 & 4.95 & 6.07  \\
& LLR-BC  & 0.58  & 0.42  & 1.78  & 0.16 & 2.82  \\
& \textbf{DREE} & 0.49  & 0.39  & 0.58  & 0.09 & 1.7\\\midrule
\multirow{5}{*}{yt-67}  & MT-Ref  & 1.36  & -  & -  & - & 1.23  \\ \cmidrule(lr){2-7} 
& FT& 8.56  & 6.75  & 15.42 & 1.81 & 4.6\\
& Li& 6.1& 0.82  & 2.89  & 5.28 & 10.5  \\
& LLR-BC  & 3.86  & 3.07  & 5.53  & 0.79 & 2.97  \\
& \textbf{DREE} & 2.48  & 1.82  & 3.94  & 0.66 & 1.58  \\ \midrule
\multirow{5}{*}{cq-144} & MT-Ref  & 1.16  & -  & -  & - & 1.32  \\ \cmidrule(lr){2-7} 
& FT& 9.74  & 8.41  & 38.88 & 1.32 & 1.32  \\
& Li& 18.71 & 13.45 & 20.58 & 5.26 & 30.01 \\
& LLR-BC  & 1.20  & 0.83  & 4.27  & 0.37 & 1.05  \\
& \textbf{DREE} & 1.26  & 0.80  & 4.75  & 0.46 & 2.15  \\
\bottomrule
\end{tabular}
\end{table*}

\section{More Experiment Results on Generated Dataset\label{app:experiment_results}}

\subsection{Performance on Seen Tasks\label{app:main_results}}

Table~\ref{tab:detailed_metric_synthetic} lists the detailed metric values of each method on each task order on the synthetic dataset. DREE is less sensitive to the task order, leading to the most stable metric values. Furthermore, DREE outperforms other methods in terms of each metric (excluding AFB and AMFB compared with Li) in most of the cases, especially the AP value.

\begin{table}[htbp]
\centering
\caption{Detailed metric values on each task order of the synthetic dataset.\label{tab:detailed_metric_synthetic}}
\begin{tabular}{c|c|c|c|c|c|c|c|c|c}
\toprule
\multirow{2}{*}{Order} & \multicolumn{1}{c|}{\multirow{2}{*}{Method}} & \multicolumn{4}{c|}{CVRP}& \multicolumn{4}{c}{TSP}\\ \cmidrule(lr){3-10}
& \multicolumn{1}{c|}{}& {AP}& {AFB}& {AMFB}  & ABPl & AP& AFB  & AMFB & ABPl \\
\midrule
\multirow{4}{*}{1} & {Fine-tuning} & {14.05} & {10.35} & {21.85} & 3.7  & 1.86 & 0.75 & 1.59 & 1.11 \\
& {Li}  & {6.12}  & {0.07}  & {0.07}  & 6.06 & 3.64 & {0.01} & {0.01} & 3.63 \\
& LLR-BC & {4.5}& {1.03}  & {1.99}  & 3.48 & 1.19 & 0.22 & 0.64 & {0.97} \\ 
& {\textbf{DREE}}& {3.16}  & {0.22}  & {0.42}  & {2.94} & {1.12} & 0.1  & 0.29 & 1.02 \\ \midrule
\multirow{4}{*}{2} & {Fine-tuning} & {5.4}& {2.6}& {8.37}  & 2.79 & 1.59 & 0.83 & 1.31 & 0.76 \\
& {Li}  & {5.62}  & {0.07}  & {0.08}  & 5.56 & 4.18 & {0.01} & {0.01} & 4.17 \\
& LLR-BC & {3.75}  & {0.91}  & {1.55}  & 2.84 & 1.18 & 0.4  & 0.51 & 0.83 \\
& {\textbf{DREE}}& {3.21}  & {0.5}& {1.03}  & {2.71} & {1.03} & 0.21 & 0.22 & {0.82 }\\ \midrule
\multirow{4}{*}{3} & {Fine-tuning} & {14.17} & {10.93} & {18.93} & 3.25 & 1.8  & 0.92 & 1.29 & 0.88 \\
& {Li}  & {6.21}  & {0.02}  & {0.03}  & 6.19 & 3.34 & {0.01} & {0.02} & 3.33 \\
& LLR-BC & {4.32}  & {1.35}  & {1.74}  & {2.97} & 1.15 & 0.31 & 0.40  & {0.84} \\
& {\textbf{DREE}}& {3.24}  & {0.24}  & {0.47}  & 3.00 & {1.02} & 0.17 & 0.19 & 0.85 \\ \midrule
\multirow{4}{*}{4} & {Fine-tuning} & {5.78}  & {2.56}  & {8.30}& 3.23 & 2.16 & 0.96 & 1.96 & 1.2  \\ 
& {Li}  & {6.04}  & {0.19}  & 0.20& 5.85 & 5.09 & {0.02} & {0.12} & 5.07 \\
& LLR-BC & {4.27}  & {1.20}& {1.68}  & 3.07 & 1.45 & 0.34 & 0.69 & 1.11 \\ 
& {\textbf{DREE}}& {2.87}  & {0.10}& {0.10}& {2.77} & {1.08} & 0.12 & 0.16 & {0.96} \\ \midrule
\multirow{4}{*}{5} & {Fine-tuning} & {4.34}  & {0.91}  & {21.28} & 3.43 & 2.44 & 0.86 & 2.06 & 1.58 \\
& {Li}  & {6.8}& {0.12}  & {0.12}  & 6.69 & 5.95 & {0.03} & {0.11} & 5.92 \\
& LLR-BC & {4.13}  & {0.71}  & {1.45}  & 3.42 & 1.94 & 0.41 & 0.99 & 1.52 \\ 
& \textbf{DREE}& {3.09}& {0.10} & {0.10} & {2.99} & {1.72} & 0.22 & 0.28 & {1.50}  \\
\midrule
\end{tabular}
\end{table}

\subsection{Generalization on Unseen Benchmark Instances\label{app:libs}}

We use the same representative instance sets as those selected in~\cite{LLR-BC,Omni} from CVRPLIB and TSPLIB. We test the solver obtained by learning from task order 1 of the synthetic CVRP and TSP dataset on CVRPLIB and TSPLIB instances, respectively. Tables~\ref{tab:cvrplib} and~\ref{tab:tsplib} report detailed results, respectively. Gaps are computed based on the reported best-known solutions. 

\begin{table*}[h]
\caption{Test results on each included CVRPLIB instance.\label{tab:cvrplib}}
\begin{adjustbox}{max width=\linewidth}
\begin{tabular}{c|c|c|c|c|c|c|c|c|c|c}
\toprule
\multirow{2}{*}{Instance} & \multicolumn{2}{c|}{MT-Ref} & \multicolumn{2}{c|}{Fine-tuning} & \multicolumn{2}{c|}{Li} & \multicolumn{2}{c|}{LLR-BC} & \multicolumn{2}{c}{\textbf{DREE}} \\
\cmidrule(lr){2-11}
& Distance & Gap (\%)& Distance& Gap (\%)& Distance & Gap (\%)& Distance   & Gap (\%)  & Distance  & Gap (\%) \\
\midrule
X-n1001-k43   & 86266.88  & 19.23  & 90196.17   
& 24.66  & 91417.62& 26.35& 94755.37& 30.96& 89635.44 & 23.88 \\
X-n101-k25& 29988.91  & 8.69   & 30540.37   & 10.69  & 30926.36& 12.09& 29478.07& 6.84 & 29539.54 & 7.06\\
X-n153-k22& 24036.21  & 13.27  & 24083.69   & 13.5   & 25059.6 & 18.09& 24751.04& 16.64& 23686.61 & 11.62 \\
X-n200-k36& 62479.91  & 6.66   & 62848.84   & 7.29   & 65328.33& 11.52& 62325.69& 6.4  & 62200.68 & 6.18\\
X-n251-k28& 41563.26  & 7.44   & 42712.81   & 10.41  & 43884.06& 13.44& 44103.78& 14.01& 41718.06 & 7.84\\
X-n303-k21& 24117.38  & 10.96  & 26153.8& 20.32  & 27413.56& 26.12& 25996.89& 19.6 & 24596.6& 13.16 \\
X-n351-k40& 29544.25  & 14.09  & 30682.58   & 18.48  & 32419.11& 25.19& 33240.53& 28.36& 29324.47 & 13.24 \\
X-n401-k29& 70920.75  & 7.21   & 71970.64   & 8.79   & 77196.08& 16.69& 81340.6 & 22.96& 70874.54 & 7.14\\
X-n459-k26& 28831.45  & 19.44  & 30027.02   & 24.39  & 32952.42& 36.51& 30024.71& 24.38& 28680.1& 18.81 \\
X-n502-k39& 75043.78  & 8.4& 74583.1& 7.74   & 77735.05& 12.29& 84633.9 & 22.26& 73331.75 & 5.93\\
X-n548-k50& 95298.19  & 9.92   & 96827.34   & 11.68  & 101873.6& 17.5 & 112970.4& 30.3 & 95291.28 & 9.91\\
X-n599-k92& 121070.5  & 11.64  & 125330.2   & 15.56  & 138531.8& 27.74& 132365.1& 22.05& 121676.1 & 12.19 \\
X-n655-k131   & 118827& 11.28  & 115713.6   & 8.37   & 124589& 16.68& 145577.2& 36.33& 115984.8 & 8.62\\
X-n701-k44& 93125.35  & 13.67  & 96309.91   & 17.56  & 95707.08& 16.83& 96237.26& 17.47& 92692.63 & 13.15 \\
X-n749-k98& 87428.02  & 13.15  & 91824.92   & 18.84  & 100277.4& 29.78& 100074  & 29.51& 89567.84 & 15.92 \\
X-n801-k40& 85760.6   & 16.98  & 87867.68   & 19.86  & 89375.21& 21.91& 179235.1& 144.49 & 84788.5& 15.66 \\
X-n856-k95& 99516.53  & 11.86  & 107203 & 20.5   & 110038.4& 23.69& 151211.4& 69.97& 105347.8 & 18.41 \\
X-n895-k37& 68330.28  & 26.87  & 69874.6& 29.73  & 70757.8 & 31.37& 97394.1 & 80.83& 67637.25 & 25.58 \\
X-n957-k87& 97976.01  & 14.64  & 104127.2   & 21.84  & 109240.4& 27.82& 130347.2& 52.52& 108971.7 & 27.5   \\
% \midrule
% Mean (Std.) & - & 12.92 (4.90) & - & 16.32 (6.39) & - & 21.66 (7.13) & - &35.57 (31.83) & - & 13.78 (6.40) \\
\bottomrule
\end{tabular}
\end{adjustbox}
\end{table*}

\begin{table*}[h]
\caption{Test results on each included TSPLIB instance.\label{tab:tsplib}}
\begin{adjustbox}{max width=\linewidth}
\begin{tabular}{c|cc|cc|cc|cc|cc}
\toprule
\multirow{2}{*}{Instance} & \multicolumn{2}{c|}{MT-Ref}  & \multicolumn{2}{c|}{Fine-tuning} & \multicolumn{2}{c|}{Li}& \multicolumn{2}{c|}{LLR-BC} & \multicolumn{2}{c}{\textbf{DREE}}   \\ \cmidrule(lr){2-11}
& {Distance} & Gap (\%) & {Distance} & Gap (\%) & {Distance} & Gap (\%) & {Distance}  & Gap (\%) & {Distance}  & Gap (\%) \\ \midrule
a280& {3042.01}  & 17.95& {3067.16}  & 18.93& {3360.59}  & 30.31& {3127.75}   & 21.28& {2944.83}   & 14.18\\
d493& {51872.83} & 48.2 & {48369.09} & 38.19& {55392.14} & 58.25& {48362.88}  & 38.17& {48186.50}  & 37.67\\
d657& {63991.38} & 30.83& {64949.99} & 32.79& {69454.31} & 42   & {64624.70}  & 32.12& {64341.84}  & 31.55\\
fl417   & {13726.64} & 15.73& {13643.5}  & 15.03& {13696.16} & 15.47& {13384.69}  & 12.85& {13487.66}  & 13.71\\
kroA100 & {22350.05} & 5.02 & {22719.97} & 6.76 & {24029.82} & 12.91& {22484.86}  & 5.65 & {22667.00}  & 6.51 \\
kroA150 & {28621.91} & 7.91 & {28467.85} & 7.33 & {30349.84} & 14.42& {28174.27}  & 6.22 & {28293.38}  & 6.67 \\
kroA200 & {32378.11} & 10.25& {32931.33} & 12.13& {34288.3}  & 16.75& {32707.89}  & 11.37& {32305.86}  & 10.00\\
kroB200 & {33546.88} & 13.96& {33121.93} & 12.52& {35114.09} & 19.29& {33433.37}  & 13.58& {32428.63}  & 10.16\\
lin318  & {49056.6}  & 16.72& {49268.01} & 17.22& {52593.37} & 25.14& {49528.87}  & 17.84& {48400.65}  & 15.16\\
p654& {42974.79} & 24.05& {41371.89} & 19.42& {45467.48} & 31.25& {41292.63}  & 19.19& {40765.16}  & 17.67\\
pcb442  & {61088.63} & 20.31& {61423.84} & 20.97& {69073.19} & 36.03& {62043.08}  & 22.18& {62127.62}  & 22.35\\
pr1002  & {353243.6} & 36.36& {347019.3} & 33.96& {390824.7} & 50.87& {353517.23} & 36.47& {345387.59} & 33.33\\
pr226   & {83418.01} & 3.79 & {83694.54} & 4.14 & {86377.18} & 7.48 & {83733.49}  & 4.19 & {84997.44}  & 5.76 \\
pr264   & {54771.85} & 11.47& {53165.7}  & 8.2  & {62884.41} & 27.98& {53237.98}  & 8.35 & {52755.43}  & 7.37 \\
pr299   & {57926.24} & 20.2 & {58615.6}  & 21.63& {62212.08} & 29.09& {57044.84}  & 18.37& {56933.31}  & 18.14\\
pr439   & {134547.6} & 25.49& {133522.1} & 24.53& {147491.2} & 37.56& {133519.80} & 24.53& {131273.62} & 22.44\\
rat575  & {8550.87}  & 26.25& {8479.65}  & 25.2 & {9702.2}   & 43.25& {8347.80}   & 23.25& {8518.76}   & 25.78\\
rat783  & {11577.88} & 31.48& {11608.1}  & 31.82& {13347.75} & 51.58& {11389.70}  & 29.34& {11447.09}  & 29.99\\
rd400   & {18396.2}  & 20.39& {18363.85} & 20.17& {19687.65} & 28.84& {18250.00}  & 19.43& {18235.69}  & 19.34\\
ts225   & {134234.4} & 5.99 & {137818.5} & 8.82 & {146137.8} & 15.39& {136657.53} & 7.91 & {132803.86} & 4.86 \\
tsp225  & {4360.97}  & 11.36& {4492.83}  & 14.73& {4593.41}  & 17.3 & {4343.73}   & 10.92& {4422.60}   & 12.94\\
u574& {46249.28} & 25.32& {48140.74} & 30.45& {51238.19} & 38.84& {46493.46}  & 25.98& {45916.29}  & 24.42\\
u724& {53539.75} & 27.75& {54104.91} & 29.1 & {60594.15} & 44.58& {53172.20}  & 26.87& {52508.60}  & 25.29\\ \bottomrule
\end{tabular}
\end{adjustbox}
\end{table*}

\subsection{More Details of Applicability Experiment\label{app:applicability_results}}

% \paragraph{To Different Base Solver.} 
We implement \textit{fine-tuning}, the most widely used strategy, LLR-BC, the state-of-the-art lifelong learning solver, and DREE on another existing cross-distribution cross-scale neural solvers, INViT~\cite{INViT} and Omni~\cite{Omni} based on their official code\footnote{https://github.com/Kasumigaoka-Utaha/INViT and https://github.com/RoyalSkye/Omni-VRP}. 
Omni is designed to produce a strong initial model that can quickly adapt to new tasks, through meta-learning. Accordingly, we adopt the meta-learned initial model provided by~\cite{Omni} as the starting model for lifelong learning with 16 batches per epoch, and a total of 1000 epochs as the setting of experiments based on POMO. 
% INViT aims at learning from one task to achieve strong zero-shot generalization to other tasks. 
For INViT, we use the same setting as the experiments with POMO as the base neural solver, i.e., 1000 epochs with 128 batches per epoch for DREE and 164 batches per epoch for other lifelong learning methods.
% It lead to a smaller per-task budget than the original setting of~\cite{INViT}.
% We expect that with a longer budget, DREE can still outperform the compared lifelong learning solvers.
Hyperparameters of the compared methods are set identically to those used on POMO. Hyperparameters of base neural solvers are set identically to those in the original papers.
Notably, we focus on comparing different lifelong learning solvers built upon the same base solver, rather than comparing across different base solvers. Results obtained with different base solvers are not directly comparable, since the training setups are different.

% \paragraph{To Periodically Stationary Secnario.} We follow the setting of~\citet{LLR-BC}

\subsection{More Details of Ablation Study\label{app:ablation}}
The ablation variant \textit{w.o.PIR} sets $\mathcal{L}_{\mathrm{PIR}}$ to 0 for all epochs throughout lifelong learning. We still perform PIR so that EE can be executed.
The ablation variant \textit{w.o.BR} removes the BR module from DREE. Consequently, EE no longer influences learning, since the enhanced behaviors are not used.
The ablation variant \textit{w.o.EE} disables behavior replacement after PIR. This modification affects the BR module (as the quality of buffered behaviors is no longer improved) but does not impact the PIR module. 
For all ablation variants, we fix $N \gets 3.56$. Because different ablations have different learning capacities, they may converge to different expected values of $M^+$. Fixing $N$ ensures that all ablation variants perform the same number of PIR executions, enabling a fairer comparison.

\subsection{More Details of Hyperparameter Analysis\label{app:hyperparameter}}

\begin{table}[htbp]
\centering
\caption{Metrics of DREE in different settings, on task order 1. 
\label{tab:parameter}
}
% \vspace{-6pt}
\begin{adjustbox}{max width=\linewidth}
 \setlength{\tabcolsep}{3.5pt}
\begin{tabular}{c|c|c|c|c|c|c|c|c|c}
\toprule
\multirow{2}{*}{Method}  & \multicolumn{4}{c|}{CVRP}  & \multirow{2}{*}{Method} & \multicolumn{4}{c}{TSP}   \\ \cmidrule(lr){2-5} \cmidrule(lr){7-10}
 & AP   & AFB  & AMFB & ABPl & & AP   & AFB  & AMFB & ABPl \\\midrule
{$\alpha$(50)} & {3.34} & {0.46} & {1.48} & {3.48} & {$\alpha$(5)} & 1.16 & 0.15 & 0.34 & 1.00 \\
$\alpha$(200)& 3.25 & 0.28 & 0.36 & 2.97 & $\alpha$(20)& 1.09 & 0.08 & 0.22 & 1.00 \\
$\beta$(0.5) & 3.27 & 0.23 & 0.79 & 3.04 & $\beta$(0.5)& 1.13 & 0.14 & 0.34 & 0.99 \\
$\beta$(2)   & 3.13 & 0.16 & 0.30 & 2.97 & $\beta$(2)  & 1.21 & 0.08 & 0.21 & 1.13 \\
$|\mathcal{B}|$(128) & 3.19 & 0.27 & 0.43 & 2.92 & $|\mathcal{B}|$(128)& 1.14 & 0.12 & 0.31 & 1.02 \\
$|\mathcal{B}|$(64)  & 3.24 & 0.29 & 0.35 & 2.96 & $|\mathcal{B}|$(64) & 1.16 & 0.14 & 0.30 & 1.02 \\
$\mathrm{UB}(8)$ & 3.30 & 0.24 & 0.63 & 3.06 & $\mathrm{UB}(8)$& 1.18 & 0.13 & 0.34 & 1.05 \\
$\mathrm{LB}(2)$ & 3.14 & 0.24 & 0.38 & 2.90 & $\mathrm{LB}(2)$& 1.11 & 0.10 & 0.26 & 1.01 \\
\textbf{DREE}& 3.16 & 0.22 & 0.42 & 2.94 & \textbf{DREE}   & 1.12 & 0.10 & 0.29 & 1.02 \\ 
\bottomrule
\end{tabular}
\end{adjustbox}

% $\alpha$(50) & 3.34 & 0.46 & 1.48 & 3.48 & 1.08 & 0.08 & 0.14 & 1.01 \\
% $\alpha$(200) & 3.25 & 0.28 & 0.36 & 2.97 & 1.37 & 0.05 & 0.07 &1.31 \\
% $\beta$(0.5) & 3.27 & 0.23 & 0.79 & 3.04 & 1.22 & 0.08 & 0.12 & 1.14\\
% $\beta$(2)  & 3.13 & 0.16 & 0.30 & 2.97 & 1.22 & 0.05 & 0.06 & 1.16\\
% $|\mathcal{B}|$(128) & 3.19 & 0.27 & 0.43 & 2.92 & 1.19 & 0.06 & 0.09 & 1.13\\
% $|\mathcal{B}|$(64) & 3.24 & 0.29 & 0.35 & 2.96 & 1.22 & 0.12 & 0.13 & 1.10\\
% $\mathrm{UB}(8)$ & 3.30 & 0.24 & 0.63 & 3.06 & 1.26 & 0.14 & 0.15 & 1.13\\
% $\mathrm{LB}(2)$ & 3.14 & 0.24 & 0.38 & 2.90 & 1.19 & 0.05 & 0.08 & 1.14\\

 % \vspace{-8pt}
\end{table}
Notably, a larger $|\mathcal{B}|$ and a smaller $N$ (with its range defined by $\mathrm{UB}$ and $\mathrm{LB}$) intuitively incur higher computational cost and may yield better performance. Therefore, we only evaluate the lower-cost variants, in which $|\mathcal{B}|$ is smaller, $\mathrm{UB}$ is larger, or $\mathrm{LB}$ is larger.

As the results listed in Table~\ref{tab:parameter} show, different hyperparameter settings lead to different performance. Arguably,  $\beta(2)$ and $\alpha(50)$ are the best tested settings for CVRP and TSP, respectively. However, compared with the performance of existing methods (cf. Table~\ref{tab:measures}), the performance changes due to different hyperparameter settings are slight, indicating the hyperparameter robustness of DREE. 

\section{Further Experiments on More Settings \label{app:further_baselie}}

\subsection{Finer-Grained Test \label{app:Fine-grained_test}}
Instead of testing only on the six principal tasks at epochs 0, 200, 400, 600, 800, and 1000, we additionally conduct tests on 41 tasks by sampling one task every 25 epochs along the 1000-task stream, with task order 1. It provides a finer-grained test and more comprehensively reflects the performance changes during the lifelong learning process. Table~\ref{tab:fine_grained_test} reports the results. Although forgetting values are slightly larger than the original results for all methods, the overall conclusions remain unchanged. The relative performance patterns across methods are consistent with those observed under the original test protocol, and the comparative conclusions remain the same. 

It supports the validity of the original principal-task evaluation. The 6 principal tasks provide an efficient evaluation protocol that faithfully reflects both absolute method performance trends and the relative comparisons among methods.

\begin{table}[htbp]
\centering
\caption{Performance on task order 1 with finer-grained test set.}
\label{tab:fine_grained_test}
% \begin{tabular}{c|cccc|cccc}
\begin{tabular}{c|c|c|c|c|c|c|c|c}
\toprule
\multirow{2}{*}{Method}& \multicolumn{4}{c|}{CVRP}& \multicolumn{4}{c}{TSP} \\ \cmidrule(lr){2-9}
& AP  & AFB  & AMFB & ABPl  & AP  & AFB  & AMFB & ABPl   
\\ \midrule
Fine-tuning&21.97 & 18.12 & 30.46 & 3.85 &2.29 & 0.85 & 1.54 & 1.44 \\
Li&7.67 & \textbf{0.06} & \textbf{0.07} & 7.60 &4.69 & \textbf{0.01} & \textbf{0.01} & 4.68\\
LLR-BC&3.96 & 0.80 & 1.12 & 3.16 & 1.68 & 0.28 & 0.65 & \textbf{1.40} \\
\textbf{DREE}&\textbf{3.10} & 0.22 & 0.33 & \textbf{2.88} & \textbf{1.55} & 0.15 & 0.26 & \textbf{1.40} \\
\bottomrule
\end{tabular}
\end{table}

\subsection{LLR-BC with Different Buffering Frequency \label{app:LLRBC_less_buffering}}
Our adaptation of LLR-BC is consistent with its original design and provides a fair comparison. LLR-BC stores buffered experiences at the final epoch of each stationary task, regardless of the number of epochs of the task. In the gradually drifting scenario, each epoch corresponds to a unique stationary task. Therefore, LLR-BC buffers at every epoch.

Nevertheless, we conduct additional experiments to analyze the effect of buffer frequency on the performance of LLR-BC. We further evaluate two LLR-BC variants that buffer every 10 epochs (denoted as LLR-BC(10)) and every 100 epochs (denoted as LLR-BC(100)), respectively, on task order 1 of the synthetic dataset. Table~\ref{tab:LLRBC_more_frequency} lists the results. Compared with the performance of DREE, the impact of buffering frequency in LLR-BC is negligible.

\begin{table}[htbp]
\centering
\caption{Performance of LLR-BC with different buffering frequency in unit of epoch.}
\label{tab:LLRBC_more_frequency}
% \begin{tabular}{c|cccc|cccc}

\begin{tabular}{c|c|c|c|c|c|c|c|c}
\toprule
\multirow{2}{*}{Method}& \multicolumn{4}{c|}{CVRP}& \multicolumn{4}{c}{TSP} \\ \cmidrule(lr){2-9}
& AP  & AFB  & AMFB & ABPl  & AP  & AFB  & AMFB & ABPl   
\\ \midrule
LLR-BC(10) & 4.62 & 1.19 & 2.01 & 3.43 & 1.17 & 1.10 & 0.22 & 1.07 \\
LLR-BC(100) & 4.69 & 1.16 & 1.78 & 3.53 & 1.19 & 0.08 & 0.16 & 1.10 \\
LLR-BC & 4.50   & 1.03  & 1.99  & 3.48 & 1.19  & 0.22   & 0.64& 0.97 \\
\bottomrule

\end{tabular}
\end{table}

\subsection{Cumulative Multi-Tasking Training \label{app:MTcumu}}
\textit{MT-cu} retrains from scratch every 200 epochs with access to all seen instances, with 200 epochs per retraining for a comparable budget to one lifelong learning run. Notably, when instance generation is uncontrollable, MT-cu incurs 500× higher storage cost than DREE, as it needs to store all seen instances. We run MT-cu on task order 1 with results listed in Table~\ref{tab:MT-cu}. 
MT-cu performs substantially worse than DREE, highlighting both the efficiency of DREE and the necessity of dedicated lifelong learning methods. 

\begin{table}[htbp]
\centering
\caption{Comparison with MT-cu.}
\label{tab:MT-cu}
% \begin{tabular}{c|cccc|cccc}
\begin{tabular}{c|c|c|c|c|c|c|c|c}
\toprule
\multirow{2}{*}{Method}& \multicolumn{4}{c|}{CVRP}& \multicolumn{4}{c}{TSP} \\ \cmidrule(lr){2-9}
& AP  & AFB  & AMFB & ABPl  & AP  & AFB  & AMFB & ABPl   
\\ \midrule
MT-cu & 4.49&0.93&5.25&3.56 & 1.25&0.04&4.43&1.21\\
\textbf{DREE} & \textbf{3.16}  & \textbf{0.22}  & \textbf{0.42}  & \textbf{2.94} &   \textbf{1.12}  & \textbf{0.10} & \textbf{0.29}& \textbf{1.02} \\
\bottomrule
\end{tabular}
\end{table}

\subsection{Continual Drift with Sudden Task changes \label{app:sudden_change}}
The synthetic dataset for the continually drifting scenario only captures smooth task transitions. Although we have already evaluated DREE under non-smooth task changes, including the Periodically Stationary scenario with abrupt shifts and real-world data exhibiting overall continual drift with substantial fluctuations, we further construct an additional variant of the continually drifting scenario that explicitly incorporates abrupt changes. This allows us to more directly assess the robustness of DREE under continual drift with sudden distributional shifts. Specifically, with task order 1, we swapped the tasks of randomly selected three pairs of epoch intervals ((23, 43)–(731, 751), (112, 162)–(517, 567), and (278, 378)–(810, 910)). For instance, the task at epoch 23 is the task originally corresponding to epoch 731. Thus, the new scenario contains abrupt task changes at the boundaries of the swapped intervals, and also exhibits recurrence of old principal-task patterns. We then run all lifelong learning methods on it, and Table~\ref{tab:measures_more_scenarios} lists the result. DREE outperforms all competing lifelong learning methods, with performance trends consistent with the smooth setting, further showing that DREE remains robust under abrupt changes and task recurrences.

\end{document}